\documentclass{article} 
\usepackage{iclr2027_conference,times}

\iclrfinalcopy

\usepackage{graphicx}
\usepackage{amsmath}

\usepackage[utf8]{inputenc} 
\usepackage[T1]{fontenc}    
\usepackage{url}            
\usepackage{booktabs}       
\usepackage{amsfonts}       
\usepackage{microtype}      
\usepackage[table,dvipsnames]{xcolor}        
\usepackage[
  colorlinks=true,
  allcolors=black,
  pdfauthor={Junyi Hu, Qiji Zhou, Lei Zhang, Yue Zhang},
  pdftitle={LAGO: Language-Guided Adaptive Object-Region Focus for Zero-Shot Visual-Text Alignment}
]{hyperref}
\usepackage{multirow}
\usepackage{adjustbox}
\usepackage{caption}
\usepackage{pifont}
\usepackage{array}
\usepackage{cleveref}
\usepackage{subcaption}
\usepackage{placeins}
\usepackage{float}
\usepackage[ruled,vlined,linesnumbered]{algorithm2e}

\DeclareCaptionFormat{longlastline}{%
  \setlength{\parfillskip}{0pt plus 0.2\linewidth}%
  #1#2#3\par}

\DontPrintSemicolon
\SetAlgoCaptionSeparator{:}
\SetAlCapNameFnt{\bfseries}
\SetAlCapFnt{\bfseries}
\SetAlgoNlRelativeSize{0}

\SetKwInput{KwIn}{Require}
\SetKwInput{KwOut}{Ensure}
\SetKw{KwRet}{return}
\SetKw{KwBreak}{break}
\SetKwProg{Fn}{Function}{}{}

\newcolumntype{C}[1]{>{\centering\arraybackslash}m{#1}}

\newcommand{\cmark}{\ding{51}}
\newcommand{\xmark}{\ding{55}}

\title{LAGO: Language-Guided Adaptive\\
Object-Region Focus for Zero-Shot\\
Visual-Text Alignment}

\author{%
  Junyi Hu$^{1,2}$\thanks{Equal contribution.}\quad
  Qiji Zhou$^{2}$\footnotemark[1]\quad
  Lei Zhang$^{3}$\quad
  Yue Zhang$^{2}$\thanks{Corresponding author.} \\
  $^{1}$Duke University \\
  $^{2}$Westlake University \\
  $^{3}$Beijing Jiaotong University \\
  \texttt{junyi.hu@duke.edu} \quad
  \texttt{zhouqiji@westlake.edu.cn} \\
  \texttt{yue.zhang@wias.org.cn}
}

\begin{document}

\maketitle
\lhead{Preprint}

\setlength{\parfillskip}{0pt plus 0.2\linewidth}
\captionsetup{format=longlastline,justification=justified,singlelinecheck=true}

\begin{abstract}
Zero-shot visual recognition often depends on localized parts, attributes, and textures that are diluted in global image representations. Existing training-free localized methods alleviate this limitation by evaluating multiple image regions, but their candidate locations are typically generated before class semantics determine where to look, restricting language to post-hoc scoring of already sampled evidence. We introduce \textbf{LAGO} (\textbf{LA}nguage-\textbf{G}uided adaptive \textbf{O}bject-region focus), which reframes localized recognition as language-guided directed region discovery. LAGO first establishes a compact, class-agnostic object-centric visual initialization, then constructs a soft semantic prototype from the intermediate prediction to actively steer subsequent search toward class-relevant evidence during inference. This object-first, language-second design addresses the circular dependence between recognizing a class and locating its supporting evidence, while preserving complementary local, contextual, and global cues. Across standard and distribution-shift benchmarks, LAGO transfers with a single fixed configuration and uses candidate regions more effectively under matched budgets. Controlled experiments further isolate the proposed mechanism, showing that allowing language to alter the search trajectory retrieves more relevant regions and improves recognition over applying the same semantic signal only after candidate generation. Code is available \href{https://github.com/JackieHuJunyi/LAGO}{here}.
\end{abstract}

\section{Introduction}
Zero-shot recognition matches images to natural-language class descriptions using a frozen vision-language model. CLIP-style methods typically rely on a global image representation, yet fine-grained decisions often depend on localized parts, attributes, or textures that are diluted at the image level~\citep{radford2021learning, menon2023visual, pratt2023platypus}. Recognition must therefore connect descriptive class cues to the regions in which their visual evidence appears. This localization is especially important when neighboring categories differ through small, spatially confined cues.

Training-free localized approaches evaluate multiple regions, but usually generate their locations through random sampling, diversity filtering, or class-agnostic saliency before text matching~\citep{li2024wca, sun2026bifta, cai2025abs}. Language can therefore score only evidence already present in the candidate pool. Enlarging an unguided pool improves coverage, but also introduces redundant or weakly relevant views. The missing capability is \emph{directed region discovery}: using class semantics to determine where visual evidence is sought, not only how sampled regions are scored.

\begin{figure}
  \centering
  \includegraphics[width=\linewidth]{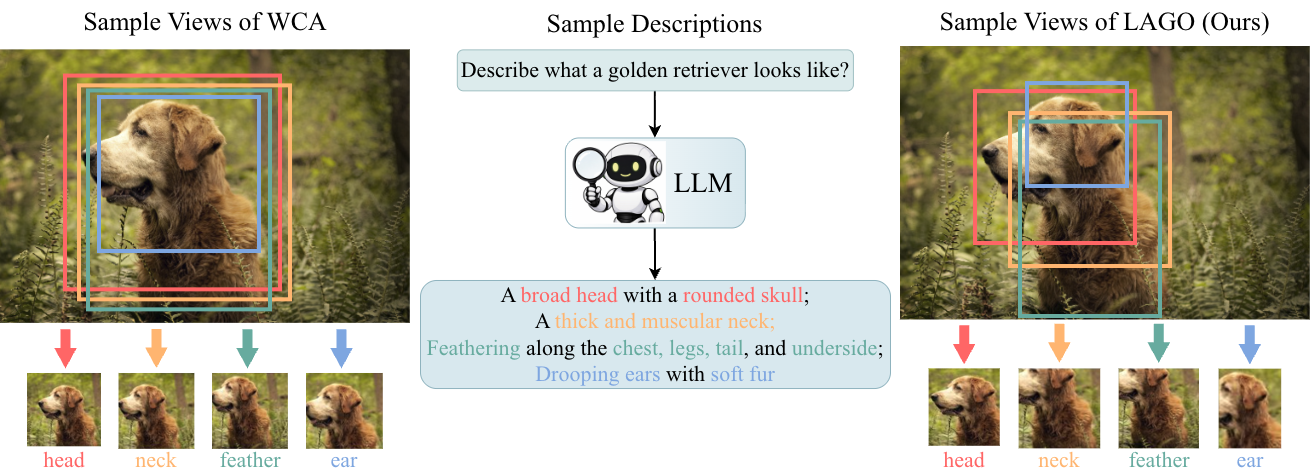}
  \caption{\textbf{From passive crop scoring to directed region discovery.}
  Given an image and candidate class (e.g., \textit{golden retriever}), fine-grained class descriptions provide semantic cues for localized recognition. WCA (left)~\citep{li2024wca} scores crops whose locations are sampled independently of class semantics. LAGO (right) first establishes an object-centric visual initialization, then uses language to actively guide subsequent search toward regions expressing the relevant attributes.}
  \label{fig:fig1}
\end{figure}

Directed discovery, however, creates a circular dependence between recognition and localization. Finding class-relevant regions requires a semantic hypothesis about the target class, while that hypothesis must itself be inferred from the visual evidence selected so far. Allowing an unreliable early prediction to guide localization too aggressively can steer the search toward evidence that reinforces the same incorrect hypothesis; removing semantic guidance entirely, on the other hand, reduces the procedure to class-agnostic search. This tension motivates an ordered strategy: first establish stable visual evidence, and only then allow semantics to direct subsequent search.

To realize this principle, we introduce \textbf{LAGO} (\textbf{LA}nguage-\textbf{G}uided adaptive \textbf{O}bject-region focus), an object-first, language-second framework for directed region discovery. As illustrated in \Cref{fig:fig1}, LAGO first performs class-agnostic object-centric search to obtain a compact and diverse visual initialization. It then constructs a soft semantic prototype from the intermediate prediction and uses region--text relevance to steer subsequent search toward class-relevant evidence. Intermediate confidence modulates the strength of semantic refinement, while the final prediction combines object-focused, contextual, and global evidence. In this way, language becomes an active search signal without requiring the initial semantic hypothesis to be treated as a reliable query.

Experiments support both the effectiveness and the mechanism of directed discovery. With one source-specified configuration frozen across target domains, LAGO transfers without target-label coefficient selection and yields positive numerical gains over full-image inference on eight datasets. Under a matched target-label-free protocol, it achieves higher accuracy than a localized baseline on four representative targets, with three differences remaining significant after paired testing and Holm correction. More importantly, a controlled intervention holds the initial proposals, Stage-1 outputs, semantic scoring, search schedule, final candidate budget, and aggregation fixed while varying whether language can influence the Stage-2 search trajectory. Directed discovery retrieves more query-aligned evidence and yields a statistically reliable recognition improvement over post-hoc semantic reranking, providing direct evidence for active semantic search beyond scoring a fixed candidate pool. Our main contributions are twofold: (i) we formulate localized zero-shot alignment as \emph{directed region discovery}, changing language from a passive scoring signal into an active guide for visual evidence acquisition; and (ii) we introduce LAGO, an object-first, language-second inference framework that realizes this principle by delaying semantic guidance until after a stable visual initialization, and validate it through target-label-free transfer, matched localized baselines, and controlled mechanism analysis.

\section{Related Work}

\noindent\textbf{Textual supervision for zero-shot recognition.}
Prompt-learning methods improve zero-shot recognition by adapting textual contexts or regularizing them for generalization~\citep{zhou2022learning, zhou2022conditional, zang2022upt, zhu2023prograd, yao2023kgcoop, khattak2023maple, khattak2023promptsrc, yao2024tcp, khattak2024protext}. Training-free approaches instead enrich class semantics with language-model-generated descriptions of parts, attributes, and broader visual concepts~\citep{menon2023visual, pratt2023platypus, roth2023waffling}. These methods improve \emph{what} semantic information is represented, but typically compare it against a single global image representation. When discriminative evidence is localized, richer descriptions alone do not determine \emph{where} that evidence should be sought. LAGO complements textual enrichment by using descriptive semantics to guide visual-region discovery.

\noindent\textbf{Localized visual-text alignment.}
Recent work incorporates regional visual-text alignment either through learned representations or through inference-time region selection~\citep{xie2025fgclip, chen2025cloc, xiao2025flair}. FG-CLIP learns global and regional alignment using detailed captions, region--caption pairs, and fine-grained negatives, while FLAIR learns text-conditioned pooling for localized image representations~\citep{xie2025fgclip, xiao2025flair}. These approaches modify or learn region-aware representations. Closer to our training-free setting, WCA reweights similarities between class descriptions and randomly sampled crops, BiFTA removes spatially overlapping views and redundant descriptions, and other methods improve class-agnostic candidate saliency before text matching~\citep{li2024wca, sun2026bifta, cai2025abs}. Despite these differences, the candidate locations in training-free pipelines are generally determined before candidate-class semantics influence where to search. LAGO instead first establishes a class-agnostic visual initialization and then allows a prediction-derived semantic prototype to alter which regions are subsequently explored and retained. The key distinction is therefore not merely regional scoring, but language-guided region discovery with a frozen vision-language model.

\noindent\textbf{Inference-time refinement and language-guided localization.}
Test-time methods adapt prompts, calibration, or view predictions to improve vision-language models under distribution shift~\citep{shu2022test, feng2023difftpt, yoon2024ctpt, samadh2023promptalign, hakim2025clipartt, zanella2024mta}. LAGO instead keeps prompts and encoder parameters fixed and changes only the visual evidence selected during inference. Language-guided grounding and referring-expression methods likewise use text to identify spatial evidence, but they assume a known query specifying the target to localize~\citep{liu2023groundingdino, xiao2024visualgroundingsurvey, yu2018mattnet, deng2021transvg, li2022glip}. Zero-shot recognition presents a different setting: the candidate label set is known, but the correct semantic query is latent. LAGO addresses this latent-query problem through an ordered object-first, language-second strategy: it establishes class-agnostic visual evidence before constructing a prediction-derived semantic prototype that guides subsequent search, with confidence modulating the strength of this guidance.

\begin{figure*}[t]
  \centering
  \includegraphics[width=\linewidth]{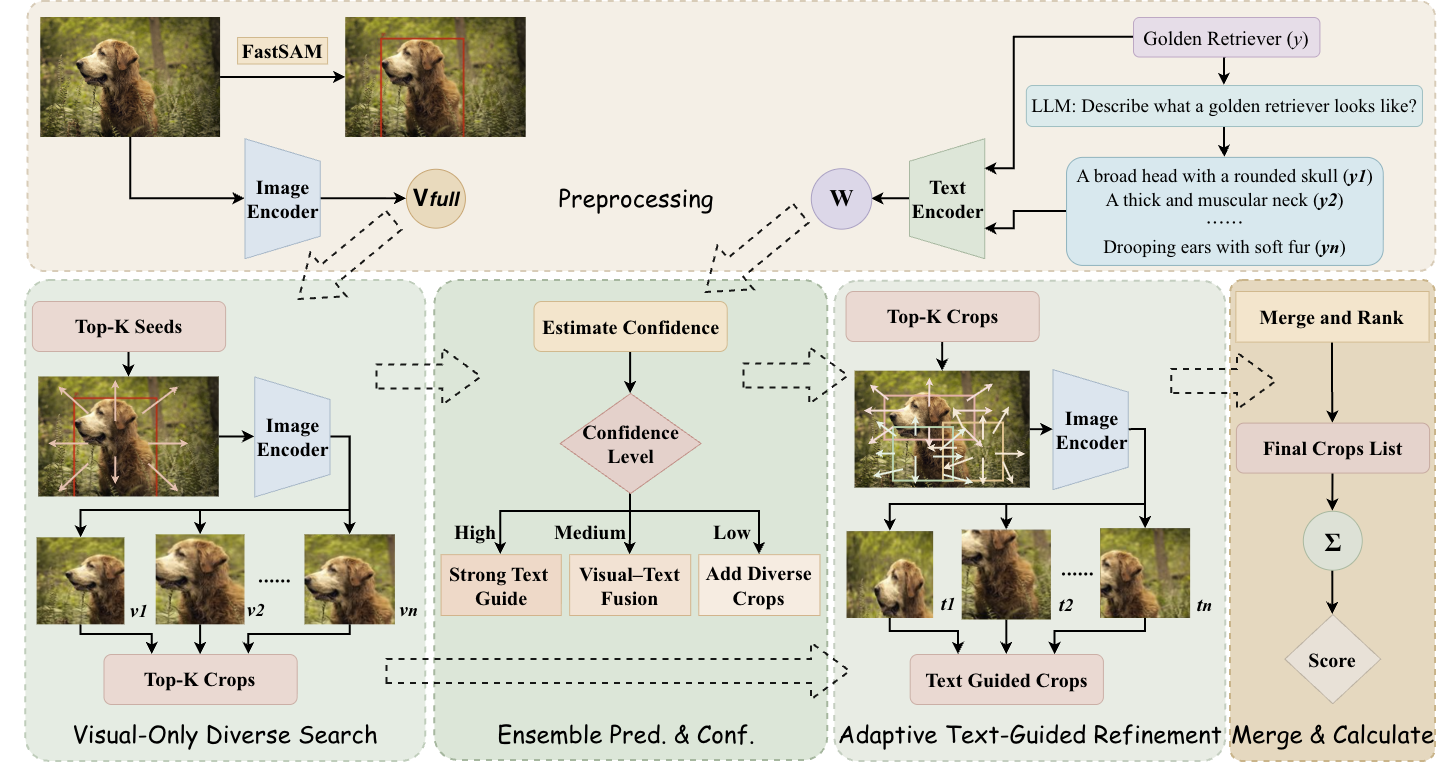}
  \caption{\textbf{Overview of LAGO.} Class-agnostic initialization first establishes a compact proposal set (\S\ref{sec:init}); a prediction-derived prototype then guides confidence-modulated Stage~2 search (\S\ref{sec:twostage}), and object-context aggregation combines local, contextual, and global evidence (\S\ref{sec:aggregation}).}
  \label{fig:lago_method}
\end{figure*}

\section{Method}

As shown in \Cref{fig:lago_method}, LAGO implements object-first, language-second region discovery. Preprocessing (\S\ref{sec:setup}) extracts full-image features, generates class-agnostic FastSAM proposals~\citep{zhao2023fastsam}, and encodes the class descriptions. Visual-only search (\S\ref{sec:init}) explores around the proposals to construct a compact, diverse region set without a class hypothesis. The resulting prediction supplies a soft text prototype, while its confidence controls how strongly that prototype guides the second search stage (\S\ref{sec:twostage}). Finally, object-context aggregation (\S\ref{sec:aggregation}) combines object-focused regions, contextual crops, and the full image for classification.

\subsection{Preliminaries}
\label{sec:setup}

Let $x$ denote an input image and let $\mathcal{Y}$ denote the candidate class set. We use a frozen CLIP model with an image encoder $E_I$ and a text encoder $E_T$~\citep{radford2021learning}. For each class $y \in \mathcal{Y}$, LAGO encodes its textual description $q_y$ as $\mathbf{t}_y = E_T(q_y) \in \mathbb{R}^d$. For each image crop $c$, it encodes the content as $\mathbf{v}(c) = E_I(c) \in \mathbb{R}^d$. Crop-level image-text alignment is measured by cosine similarity,
\begin{equation}
s(c,y) = \mathrm{sim}\bigl(\mathbf{v}(c), \mathbf{t}_y\bigr).
\end{equation}

\subsection{Class-Agnostic Object-Centric Candidate Initialization}
\label{sec:init}

At the beginning of zero-shot inference, the target class and its relevant semantics are unknown. Allowing an early prediction to control localization can bias subsequent region selection toward misleading visual evidence. LAGO therefore begins with \emph{class-agnostic} candidate initialization.

For each image, we obtain a small set of object-centric proposal boxes,
\begin{equation}
\mathcal{B}_0 = \{b^{(0)}_1, b^{(0)}_2, \dots, b^{(0)}_M\},
\end{equation}
using FastSAM as an off-the-shelf, class-agnostic proposal generator~\citep{zhao2023fastsam}. These proposals provide compact object-centric seeds for the visual-only search in \S\ref{sec:twostage}, restricting subsequent exploration to plausible visual regions before any class hypothesis is introduced. This initialization reduces the search space while preserving diverse object-level coverage for later semantic refinement. The complete proposal filtering and search implementation are provided in Appendix~\ref{app:search}.

\subsection{Confidence-Aware Two-Stage Region Discovery}
\label{sec:twostage}

Directed region discovery creates a circular dependence between recognition and localization: identifying class-relevant regions requires a semantic hypothesis, while that hypothesis must itself be inferred from the visual evidence selected so far. An unreliable early hypothesis can therefore steer the search toward evidence that reinforces the same error. LAGO addresses this dependence through an ordered two-stage procedure: Stage~1 establishes class-agnostic visual evidence, and only then does Stage~2 allow the intermediate semantic hypothesis to influence subsequent region search.

\paragraph{Scoring functions.}
For each candidate box $b$, LAGO uses a visual score and a text score:
\begin{equation}
S_{\mathrm{visual}}(b) = p(b),
\end{equation}
\begin{equation}
S_{\mathrm{text}}(b) = q(b)
= \mathrm{sim}\!\bigl(\mathbf{v}(b), \mathbf{w}_{\mathrm{text}}\bigr).
\end{equation}
Here, $S_{\mathrm{visual}}(b)$ measures class-agnostic crop quality, whereas $S_{\mathrm{text}}(b)$ measures semantic relevance to the prediction-derived prototype $\mathbf{w}_{\mathrm{text}}$, which becomes available after Stage~1. Both scores are later reused in crop aggregation during final evidence fusion.

\paragraph{Stage 1: visual-only discovery.}
Starting from the object-centric proposal seeds $\mathcal{B}_0$, Stage~1 performs class-agnostic local search using only visual evidence. For each explored box $b$,
\begin{equation}
S^{(1)}(b) = S_{\mathrm{visual}}(b),
\end{equation}
which sets the semantic-guidance weight to zero. The candidate set, denoted by $\mathcal{C}^{(1)}$, provides a compact visual initialization without relying on an uncertain class hypothesis.

Using the top-ranked regions in $\mathcal{C}^{(1)}$, we compute an intermediate crop-based prediction with logits $\mathbf{z}^{(1)} \in \mathbb{R}^{|\mathcal{Y}|}$. These logits serve two distinct roles: they produce a prediction-derived semantic prototype $\mathbf{w}_{\mathrm{text}}$ for subsequent search and an intermediate confidence estimate
\begin{equation}
c = \phi\bigl(\mathbf{z}^{(1)}\bigr),
\end{equation}
where $\phi(\cdot)$ denotes the confidence estimator. The prototype determines the semantic direction of Stage~2, while confidence controls how strongly that direction influences the search.

\paragraph{Stage 2: confidence-modulated semantic discovery.}
Stage~2 introduces semantic relevance into the search objective, allowing language to change which candidate locations are subsequently explored and retained. The sample-specific confidence determines the semantic-guidance strength,
\begin{equation}
\gamma = \psi(c),
\end{equation}
where $\psi(\cdot)$ maps confidence to a guidance weight. The Stage~2 score is
\begin{equation}
S^{(2)}(b)
=
(1-\gamma)\,S_{\mathrm{visual}}(b)
+
\gamma\,S_{\mathrm{text}}(b).
\end{equation}

High-confidence samples therefore receive stronger semantic guidance, whereas low-confidence samples remain closer to the class-agnostic visual objective. Crucially, $S^{(2)}$ is used during search rather than only after a fixed crop pool has been generated, so the semantic prototype can alter the search trajectory itself. This object-first, language-second ordering makes language an active region-discovery signal while limiting the influence of unreliable early hypotheses.

Algorithm~\ref{alg:two_stage} summarizes the complete two-stage inference procedure. The proposal-centered local-search operator is detailed in Appendix~\ref{app:search} and Algorithm~\ref{alg:search_topk}.

\begin{algorithm}[t]
\caption{Confidence-Aware Two-Stage Region Discovery}
\label{alg:two_stage}
\KwIn{image $x$, proposal set $\mathcal{B}_0$, full-image feature $\mathbf{v}_{\mathrm{full}}$, class text bank $\mathbf{W}$}
\KwOut{final crop set $\mathcal{C}$, semantic prototype $\mathbf{w}_{\mathrm{text}}$, and confidence estimate $c$}

\Fn{\textnormal{\textsc{TwoStageRegionDiscovery}}$(x,\mathcal{B}_0,\mathbf{v}_{\mathrm{full}},\mathbf{W})$}{

    \tcp{Stage 1: class-agnostic visual discovery}
    $\mathcal{C}^{(1)} \gets 
    \bigcup_{b \in \mathcal{B}_0}
    \textsc{SearchTopKCrops}
    (x,b,\mathbf{v}_{\mathrm{full}},\texttt{None},0)$\;

    \tcp{Intermediate semantic hypothesis and confidence}
    $\mathbf{v}_{\mathrm{ens}} 
    \gets \textsc{MeanPoolTopCrops}(\mathcal{C}^{(1)})$\;

    $\mathbf{z}^{(1)} 
    \gets \mathbf{v}_{\mathrm{ens}}^{\top}\mathbf{W}$\;

    $(\hat{y}^{(1)}, c) 
    \gets \textsc{PredictionConfidence}(\mathbf{z}^{(1)})$\;

    $\mathbf{w}_{\mathrm{text}} 
    \gets \textsc{ConstructTextPrototype}(\mathbf{z}^{(1)},\mathbf{W})$\;

    $\gamma \gets \psi(c)$\;

    \tcp{Stage 2: confidence-modulated semantic discovery}
    $\mathcal{C}^{(2)} \gets 
    \bigcup_{b \in \textsc{TopBoxes}(\mathcal{C}^{(1)})}
    \textsc{SearchTopKCrops}
    (x,b,\mathbf{v}_{\mathrm{full}},\mathbf{w}_{\mathrm{text}},\gamma)$\;

    \tcp{Merge and rank discovered regions}
    $\mathcal{C} 
    \gets \textsc{MergeAndRefine}
    (\mathcal{C}^{(1)},\mathcal{C}^{(2)},x,\mathbf{w}_{\mathrm{text}},\gamma)$\;

    $\mathcal{C} 
    \gets \textsc{RankCrops}(\mathcal{C})$\;

    \KwRet{$\mathcal{C},\mathbf{w}_{\mathrm{text}},c$}\;
}
\end{algorithm}

\subsection{Object-Context Dual-Channel Aggregation}
\label{sec:aggregation}

Localized alignment improves recognition, but object crops alone may miss context~\citep{yu2023zeroshotris}. LAGO therefore aggregates object, context, and full-image evidence through the exact dual-channel computation used in all reported experiments and final predictions.

After region discovery, valid crops are divided into an object channel $\mathcal{C}_o$ and a context channel $\mathcal{C}_c$, while the full image contributes logits $\mathbf{z}_{\mathrm{full}}$. For each crop $c_i$, the visual and text scores from Sec.~\ref{sec:twostage} are mixed into one ranking score before any within-channel normalization is applied:
\begin{equation}
 r_i=(1-\beta)p(c_i)+\beta q(c_i), \qquad
 q(c_i)=\mathrm{sim}\!\bigl(\mathbf{v}(c_i),\mathbf{w}_{\mathrm{text}}\bigr).
\end{equation}
Within each channel $\kappa\in\{o,c\}$, we normalize the mixed score and aggregate crop features using the implementation temperature scale $s_a$ to obtain the final channel representation:
\begin{equation}
 w_i^{(\kappa)}=\operatorname{softmax}_{i\in\mathcal{C}_\kappa}(s_a r_i),
 \qquad
 \mathbf{v}_{\kappa}=\operatorname{norm}\!\left(\sum_{i\in\mathcal{C}_\kappa}w_i^{(\kappa)}\mathbf{v}(c_i)\right).
\end{equation}
The normalized object and context features are fused before classification with the text bank $\mathbf{W}$:
\begin{equation}
 \mathbf{v}_{\mathrm{dc}}=\operatorname{norm}\!\left(\alpha_{\mathrm{dc}}\mathbf{v}_{o}+(1-\alpha_{\mathrm{dc}})\mathbf{v}_{c}\right),
 \qquad
 \mathbf{z}_{\mathrm{dc}}=\mathbf{v}_{\mathrm{dc}}^{\top}\mathbf{W},
\end{equation}
after which the dual-channel logits are interpolated with the full-image logits. Here $\lambda$ is the full-image weight, exactly matching the evaluated implementation and Appendix~\ref{app:confidence_mapping}:
\begin{equation}
\mathbf{z}_{\mathrm{final}}=
\lambda\,\mathbf{z}_{\mathrm{full}}+(1-\lambda)\,\mathbf{z}_{\mathrm{dc}}.
\end{equation}

\section{Experiments}

We evaluate LAGO on standard zero-shot recognition and natural distribution-shift benchmarks, together with controlled analyses of its region-discovery mechanism.

\subsection{Experimental Setup}

\paragraph{Datasets.}
We evaluate on six standard benchmarks: ImageNet~\cite{deng2009imagenet}, CUB~\cite{welinder2010cub}, Oxford Pets~\cite{parkhi2012oxfordpets}, DTD~\cite{cimpoi2014dtd}, Food101~\cite{bossard2014food101}, and Places365~\cite{zhou2017places365}, covering generic, fine-grained, texture, food, and scene recognition across diverse domains. We further evaluate natural distribution shift on ImageNet-V2~\cite{recht2019imagenetv2}, ImageNet-R~\cite{hendrycks2021imagenetr}, ImageNet-S~\cite{wang2019imagenetsketch}, and ImageNet-A~\cite{hendrycks2021imageneta} under varied conditions.

\paragraph{Baselines.}
For standard zero-shot recognition, we compare with CLIP~\cite{radford2021learning}, CLIP-E~\cite{radford2021learning}, CuPL~\cite{pratt2023platypus}, BiFTA~\cite{sun2026bifta}, and WCA~\cite{li2024wca}, covering hand-crafted prompting, language-model-based text enrichment, and localized visual-text alignment. For natural-shift evaluation, we additionally compare with CoOp~\cite{zhou2022learning}, CoCoOp~\cite{zhou2022conditional}, UPT~\cite{zang2022upt}, ProGrad~\cite{zhu2023prograd}, KgCoOp~\cite{yao2023kgcoop}, TPT~\cite{shu2022test}, and DiffTPT~\cite{feng2023difftpt}.

\paragraph{Implementation and evaluation protocols.}
LAGO uses frozen CLIP ViT-B/32, ViT-B/16, and ViT-L/14 backbones~\cite{radford2021learning}, with no encoder updates. Unless stated otherwise, it uses proposal-centered initialization, confidence-modulated two-stage discovery, IoU-diverse filtering, and object-context aggregation. The diagnostic LAGO rows in Tables~\ref{tab:main_results}--\ref{tab:natural_shift} select $(\beta,\alpha_{\mathrm{dc}},\lambda)$ from 330 score-level configurations using labels from the corresponding evaluation cache and are therefore \emph{evaluation-label-calibrated}, rather than strict target-label-free results. We report fixed target-label-free transfer and controlled protocol variants separately, and compare absolute accuracies only within matched evaluation settings. Complete settings and protocol controls are provided in Appendix~\ref{app:method_details}. All experiments use a single NVIDIA A100 GPU.

\paragraph{Text descriptions.}
Following prior description-based zero-shot methods~\citep{menon2023visual, pratt2023platypus, sun2026bifta}, we use offline class descriptions throughout inference; their construction and optional prototype reweighting are detailed in Appendix~\ref{app:text}.

\subsection{Benchmark Comparisons and Protocol Controls}

\begin{table}[!t]
\centering
\caption{\textbf{Paired target-label-free comparison with a localized baseline.}
WCA uses 120 random views; fixed LAGO\@120 uses 120 candidate-region slots plus a separately encoded full image. $\Delta$ denotes LAGO minus WCA, and intervals are paired $95\%$ bootstrap CIs. $\dagger$ indicates significance after exact McNemar testing and Holm correction across the four comparisons.}
\label{tab:fixed_wca_transfer}
\scriptsize
\setlength{\tabcolsep}{3pt}
\begin{adjustbox}{max width=\linewidth,center}
\begin{tabular}{l c r r c c c c}
\toprule
\textbf{Dataset ($n$)} & \textbf{Backbone} & \textbf{WCA@120} & \textbf{LAGO@120} & $\boldsymbol{\Delta}$ \textbf{[95\% CI]} & \textbf{$b/c$} & \textbf{Exact $p$} & \textbf{Holm $p$} \\
\midrule
CUB (5,794) & ViT-L/14 & 63.72 & \textbf{64.48} & \textcolor{ForestGreen}{+0.76 $[+0.13,+1.38]$}$^\dagger$ & 193/149 & $1.99{\times}10^{-2}$ & $3.99{\times}10^{-2}$ \\
DTD (1,880) & ViT-L/14 & 61.70 & \textbf{62.34} & \textcolor{ForestGreen}{+0.64 $[-0.21,+1.49]$} & 40/28 & $0.182$ & $0.182$ \\
Places365 (36,500) & ViT-B/32 & 40.66 & \textbf{44.32} & \textcolor{ForestGreen}{+3.66 $[+3.45,+3.87]$}$^\dagger$ & 1442/107 & $4.55{\times}10^{-299}$ & $1.82{\times}10^{-298}$ \\
ImageNet-A (7,500) & ViT-B/16 & 55.15 & \textbf{57.95} & \textcolor{ForestGreen}{+2.80 $[+2.09,+3.51]$}$^\dagger$ & 471/261 & $7.37{\times}10^{-15}$ & $2.21{\times}10^{-14}$ \\
\bottomrule
\end{tabular}
\end{adjustbox}
\end{table}

Our primary controlled comparison uses a fixed, target-label-free protocol against the closest localized baseline. Both methods evaluate 120 candidate regions per image using matched evaluation samples, labels, text descriptions, and frozen backbones; LAGO additionally fuses a separately encoded full image. We align predictions by sample, compute 100,000 paired-image percentile bootstrap replicates (seed 0), apply two-sided exact McNemar tests, and correct the four $p$-values using Holm's method.

Table~\ref{tab:fixed_wca_transfer} shows higher LAGO accuracy on all four targets; the CUB, Places365, and ImageNet-A differences remain significant after Holm correction, while DTD does not. Appendix~\ref{app:fixed_transfer} reports the frozen transfer tuple, full-image controls, view-budget audit, and a two-backbone DTD evaluation.

\begin{table*}[!t]
\centering
\captionsetup{skip=8pt}
\caption{\textbf{Benchmark accuracy (\%) under dataset-specific operating-point selection across CLIP backbones.}
LAGO selects fusion coefficients separately for each dataset--backbone pair, while baseline rows follow their reported tuned settings. This table evaluates tuned benchmark performance; fixed target-label-free comparisons are reported separately in Table~\ref{tab:fixed_wca_transfer}.}
\label{tab:main_results}

\footnotesize
\setlength{\tabcolsep}{3.6pt}
\renewcommand{\arraystretch}{1.15}

\begin{adjustbox}{max width=\textwidth,center}
\begin{tabular}{lcccccccccccccccccc}
\toprule

\multirow{2}{*}{Method}
& \multicolumn{3}{c}{ImageNet}
& \multicolumn{3}{c}{CUB}
& \multicolumn{3}{c}{Oxford Pets}
& \multicolumn{3}{c}{DTD}
& \multicolumn{3}{c}{Food101}
& \multicolumn{3}{c}{Places365} \\

\cmidrule(lr){2-4}
\cmidrule(lr){5-7}
\cmidrule(lr){8-10}
\cmidrule(lr){11-13}
\cmidrule(lr){14-16}
\cmidrule(lr){17-19}

& B/32 & B/16 & L/14
& B/32 & B/16 & L/14
& B/32 & B/16 & L/14
& B/32 & B/16 & L/14
& B/32 & B/16 & L/14
& B/32 & B/16 & L/14 \\

\midrule

CLIP~\cite{radford2021learning}
& 62.05 & 66.74 & 73.48
& 51.21 & 56.01 & 62.12
& 85.04 & 88.14 & 93.24
& 42.93 & 42.98 & 52.61
& 82.60 & 88.40 & 92.55
& 38.51 & 39.27 & 39.63 \\

CLIP-E~\cite{radford2021learning}
& 63.37 & 68.37 & 75.52
& 52.74 & 56.16 & 62.53
& 87.38 & 89.10 & 93.62
& 43.83 & 45.27 & 55.43
& 83.93 & 88.83 & 93.07
& 39.28 & 40.30 & 40.55 \\

CLIP-D~\cite{menon2023visual}
& 63.01 & 68.04 & 75.03
& 52.69 & 57.08 & 63.26
& 84.46 & 87.52 & 93.30
& 44.20 & 46.17 & 55.05
& 84.12 & 88.85 & 93.03
& 39.90 & 40.34 & 40.55 \\

Waffle~\cite{roth2023waffling}
& 63.30 & 68.12 & 75.31
& 52.04 & 56.89 & 62.27
& 85.50 & 86.51 & 91.55
& 42.98 & 44.68 & 54.31
& 83.98 & 89.06 & 93.33
& 39.47 & 40.76 & 40.89 \\

CuPL~\cite{pratt2023platypus}
& 64.37 & 69.61 & 76.62
& 49.76 & 56.42 & 62.15
& 87.03 & 91.14 & 94.33
& 47.50 & 50.53 & 60.59
& 84.20 & 88.98 & 93.37
& 39.08 & 39.83 & 40.77 \\

BiFTA~\cite{sun2026bifta}
& 66.83 & 71.14 & 77.82
& 58.24 & 60.06 & 65.67
& 89.74 & 91.67 & 94.96
& 53.22 & 54.64 & 62.45
& 86.43 & 90.11 & 93.97
& 41.55 & 42.12 & 42.98 \\

WCA~\cite{li2024wca}
& 66.84 & 71.08 & 77.32
& 56.91 & 59.78 & 65.24
& 89.89 & 92.23 & 94.66
& 49.39 & 52.79 & 61.78
& 86.40 & 90.01 & 93.96
& 40.66 & 41.43 & 42.23 \\

\midrule
\rowcolor{gray!10}
\textbf{Ours}
& \textbf{67.98} & \textbf{71.72} & \textbf{77.97}
& \textbf{59.16} & \textbf{62.97} & \textbf{66.23}
& \textbf{90.71} & \textbf{92.78} & \textbf{95.07}
& \textbf{53.49} & \textbf{55.21} & \textbf{63.90}
& \textbf{86.96} & \textbf{90.54} & \textbf{94.35}
& \textbf{44.32} & \textbf{45.56} & \textbf{47.28} \\

\rowcolor{gray!5}
$\Delta$
& \textcolor{ForestGreen}{+1.14}
& \textcolor{ForestGreen}{+0.64}
& \textcolor{ForestGreen}{+0.65}
& \textcolor{ForestGreen}{+2.25}
& \textcolor{ForestGreen}{+3.19}
& \textcolor{ForestGreen}{+0.99}
& \textcolor{ForestGreen}{+0.82}
& \textcolor{ForestGreen}{+0.55}
& \textcolor{ForestGreen}{+0.41}
& \textcolor{ForestGreen}{+4.10}
& \textcolor{ForestGreen}{+2.42}
& \textcolor{ForestGreen}{+2.12}
& \textcolor{ForestGreen}{+0.56}
& \textcolor{ForestGreen}{+0.53}
& \textcolor{ForestGreen}{+0.39}
& \textcolor{ForestGreen}{+3.66}
& \textcolor{ForestGreen}{+4.13}
& \textcolor{ForestGreen}{+5.05} \\

\bottomrule
\end{tabular}
\end{adjustbox}
\end{table*}

Table~\ref{tab:main_results} compares dataset-specific tuned operating points across methods. Under this protocol, LAGO attains the highest accuracy in every dataset--backbone column across all three CLIP backbones. The gains over WCA are particularly pronounced on CUB, DTD, and Places365, reaching up to \textbf{+3.19}, \textbf{+4.10}, and \textbf{+5.05} points, respectively. Improvements on ImageNet, Oxford Pets, and Food101 are smaller but remain positive, indicating that the benefit extends beyond localized-evidence-dominated tasks across diverse recognition settings.

To separate tuned benchmark performance from target-label-free transfer, we additionally report fixed-protocol controls. On ImageNet-V2, the source-specified configuration reaches $65.45\%$; held-out calibration differs from the same tuple by only $0.02$ points on a shared report set, while the full-test oracle is $0.24$ points higher. Freezing this tuple before target evaluation yields positive numerical gains over full-image inference on all eight transfer datasets (Appendix~\ref{app:fixed_transfer}). Together with the matched comparison in Table~\ref{tab:fixed_wca_transfer}, these results support broad transfer across diverse target domains without relying on target-domain coefficient selection.

\begin{table*}[!t]
\centering
\captionsetup{skip=8pt}

\caption{\textbf{Benchmark accuracy (\%) under natural distribution shifts.}
TP, VP, TTP, and LLM denote textual prompting, visual prompting, test-time prompting, and large language models, respectively. ``Tuned'' indicates whether the model is fine-tuned on ImageNet; ``Source'' and ``Target'' denote in-distribution and distribution-shift performance. LAGO uses dataset-specific score-level operating-point selection, while baseline rows follow their reported evaluation settings. Fixed target-label-free comparisons are reported separately in Table~\ref{tab:fixed_wca_transfer} and Appendix~\ref{app:fixed_transfer}.}

\label{tab:natural_shift}

\footnotesize
\setlength{\tabcolsep}{5pt}
\renewcommand{\arraystretch}{1.2}

\begin{adjustbox}{max width=\textwidth,center}

\begin{tabular}{C{4.5cm}cccccccc}
\toprule

\multirow{2}{*}{Method}
& \multirow{2}{*}{Prompts}
& \multirow{2}{*}{Tuned?}
& \multicolumn{1}{c}{Source}
& \multicolumn{4}{c}{Target}
& \multirow{2}{*}{Average}

\\

\cmidrule(lr){4-4}
\cmidrule(lr){5-8}

&
&
& ImageNet
& ImageNet-V2
& ImageNet-R
& ImageNet-S
& ImageNet-A
&

\\

\midrule

CoOp~\cite{zhou2022learning}              & TP            & \cmark & 71.51 & 64.20 & 75.21 & 47.99 & 49.71 & 61.72 \\
CoCoOp~\cite{zhou2022conditional}         & VP+TP         & \cmark & 71.02 & 64.07 & 76.18 & 48.75 & 50.63 & 62.13 \\
UPT~\cite{zang2022upt}                    & VP+TP         & \cmark & \textbf{72.63} & 64.35 & 76.24 & 48.66 & 50.66 & 62.51 \\
ProGrad~\cite{zhu2023prograd}             & TP            & \cmark & \underline{72.24} & 64.73 & 74.58 & 47.99 & 49.39 & 61.79 \\
KgCoOp~\cite{yao2023kgcoop}               & TP            & \cmark & 71.20 & 64.10 & 76.70 & 48.97 & 50.69 & 62.33 \\

\midrule

TPT~\cite{shu2022test}                    & TTP           & \cmark & 69.70 & 64.30 & 73.90 & 46.40 & 53.67 & 61.59 \\
DiffTPT~\cite{feng2023difftpt}            & TTP           & \cmark & 70.30 & \underline{65.10} & 75.00 & 46.80 & 55.68 & 62.58 \\

\midrule

CLIP~\cite{radford2021learning}           & Hand-crafted  & \xmark & 66.74 & 60.83 & 73.96 & 46.15 & 47.77 & 59.09 \\
CLIP-E~\cite{radford2021learning}         & Hand-crafted  & \xmark & 68.37 & 61.90 & 77.40 & 47.87 & 49.00 & 60.91 \\
CuPL~\cite{pratt2023platypus}             & LLM-TP        & \xmark & 69.61 & 63.27 & 77.10 & 48.80 & 50.77 & 61.91 \\
WCA~\cite{li2024wca}                      & LLM-TP+VP     & \xmark & 71.08 & 64.71 & \underline{78.06} & \underline{50.18} & \underline{56.13} & \underline{64.03} \\

\midrule
\rowcolor{gray!10}

\textbf{Ours}
& \textbf{LLM-TP+VP}
& \xmark
& 71.72
& \textbf{65.90}
& \textbf{78.91}
& \textbf{50.81}
& \textbf{59.16}
& \textbf{65.30}

\\

\bottomrule
\end{tabular}

\end{adjustbox}
\end{table*}

Under the dataset-specific tuned protocol in Table~\ref{tab:natural_shift}, LAGO improves over WCA by \textbf{+1.19}, \textbf{+0.85}, \textbf{+0.63}, and \textbf{+3.03} points on ImageNet-V2, ImageNet-R, ImageNet-S, and ImageNet-A, respectively. Including the ImageNet source domain, this yields a \textbf{+1.27} point gain in average accuracy without encoder updates. We separately assess target-label-free robustness in Appendix~\ref{app:fixed_transfer}. With the fusion tuple frozen before target evaluation, LAGO improves over full-image inference by \textbf{+7.01} points on ImageNet-A and \textbf{+0.20} points on ImageNet-R. This separation distinguishes tuned benchmark performance from fixed cross-domain transfer.

Beyond accuracy, we evaluate \emph{candidate efficiency} under matched candidate budgets. Proposal-centered initialization and two-stage discovery prioritize informative and diverse regions, enabling LAGO to extract more useful visual evidence from a limited candidate budget. Across most operating points, LAGO achieves stronger recognition accuracy than random-view sampling under the same candidate budget, demonstrating more effective use of candidate evaluations (Appendix~\ref{app:runtime}).

\begin{figure}[t]
    \centering
    \includegraphics[width=0.90\linewidth]{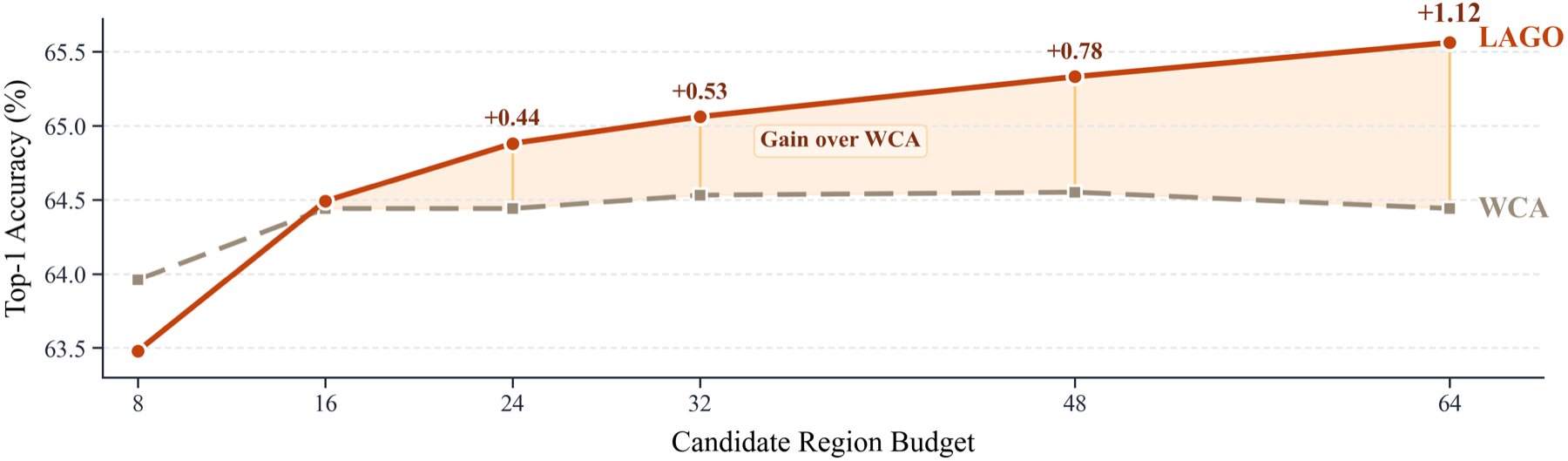}
    \caption{\textbf{Candidate-region budget analysis of LAGO.}
On ImageNet-V2, LAGO becomes stronger than WCA once the matched region budget reaches 16, and the advantage increases with additional candidate regions. Results are from a single matched budget sweep; separate resampling and deployment analyses are reported in Appendix~\ref{app:resampling}--\ref{app:runtime}.}
    \label{fig:ablation_efficiency}
\end{figure}

\subsection{Analysis: Where Do the Gains Come From?}

\paragraph{Candidate-region efficiency.}
Figure~\ref{fig:ablation_efficiency} shows that LAGO is slightly weaker than WCA at the smallest budget of 8, but becomes stronger from 16 regions onward, with the gap widening to \textbf{+1.12} points at 64 regions. This transition is consistent with the intended mechanism: proposal-centered initialization first establishes viable object coverage, after which directed semantic discovery uses the additional budget to retrieve increasingly informative regions. Full candidate-budget settings and numerical comparisons are provided in Appendix~\ref{app:efficiency}.

\paragraph{Directed discovery versus post-hoc reranking.}
We compare directed discovery with post-hoc reranking on a fixed 1,000-image CUB subset using frozen CLIP ViT-B/16. Both variants share proposals, Stage~1 outputs, search schedule, 120 final slots, and aggregation; only directed discovery uses semantic relevance to choose the next Stage~2 neighborhood. It changes 786 crop sets, increases Query Top-3 cosine by $0.00043$, and improves accuracy from $60.60\%$ to $61.40\%$ ($+0.80$ points, paired $95\%$ CI $[+0.20,+1.50]$, exact McNemar $p=0.0215$). These results support language-guided search beyond post-hoc reranking. Appendix~\ref{app:directed_discovery_control} provides full protocol details and controls.

\paragraph{Confidence-aware behavior.}
Among cases corrected by adaptive but not fixed guidance, $77/84$ ($91.7\%$) are low-confidence and $7/84$ ($8.3\%$) are medium-confidence. These corrections therefore concentrate in the regime where the controller is intended to be most active. In the fixed audit, adaptive guidance is $0.25$ points above fixed guidance, but no component difference remains significant after Holm correction. We therefore interpret confidence as a behaviorally meaningful sample-level control signal, while the ordered two-stage framework provides the stronger evidence for the overall gain.

\paragraph{Stage-wise behavior.}
Stage~1 identifies salient object-centric regions without class semantics, whereas Stage~2 uses the intermediate prediction to shift attention toward class-relevant evidence. This progression visualizes the ordered design: class-agnostic localization first establishes stable visual evidence, after which semantic guidance directs subsequent refinement. Appendix~\ref{app:stage_vis} provides representative examples across datasets for qualitative comparison.

\paragraph{Failure cases.}
In the analyzed ImageNet-R failure set, semantic mismatch is the largest observed category ($50.0\%$), followed by cluttered or poorly localized regions ($21.4\%$), other cases ($21.4\%$), and visually ambiguous classes ($7.1\%$). LAGO remains particularly vulnerable when text cues are misleading, proposals miss discriminative regions, or competing categories share similar local attributes. Representative examples of all four failure categories under natural distribution shift are provided in Appendix~\ref{app:failure}. These cases distinguish text-side errors from localization and visual-ambiguity failures, showing that stronger region discovery alone cannot resolve every error under distribution shift.

\begin{figure}[t]
    \centering
    \includegraphics[width=\linewidth]{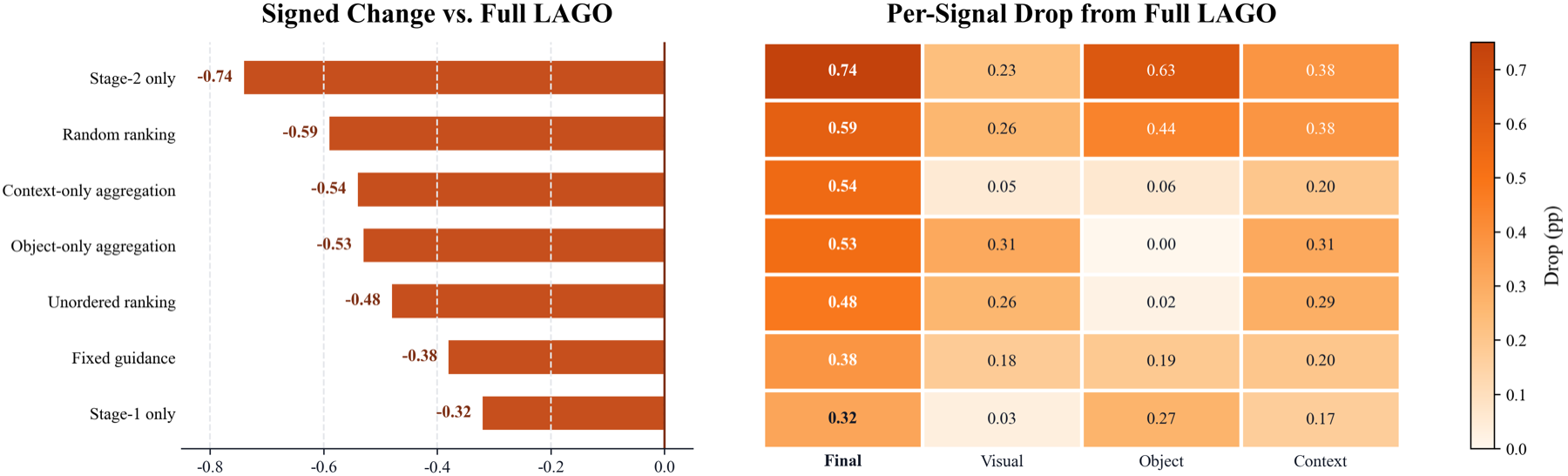}
    \caption{\textbf{Accuracy-oriented component ablations of LAGO.}
Values denote gains of Full LAGO over simplified inference variants under the calibrated accuracy-oriented setting.}
    \label{fig:ablation_effectiveness}
\end{figure}

\subsection{Component Ablations}

We ablate three aspects of LAGO: the two-stage design, the guidance and candidate-selection policy, and object-context aggregation (Figure~\ref{fig:ablation_effectiveness}). These comparisons examine the roles of delayed semantic conditioning, adaptive region selection, and complementary evidence aggregation.

\paragraph{Effect of the two-stage design.}
In the accuracy-oriented configuration, Full LAGO exceeds the Stage~1-only and Stage~2-only variants by \textbf{0.32} and \textbf{0.74} points, respectively. Stage~1 lacks semantic refinement, whereas Stage~2 applies language guidance without first establishing stable visual evidence. Together with the fixed-setting audit, this pattern supports delayed semantic conditioning and the ordered two-stage design. Appendix~\ref{app:stage2_only_cases} provides representative Stage~2-only failures.

\paragraph{Effect of guidance and region selection.}
Fixed guidance is \textbf{0.38} points below adaptive guidance in the accuracy-oriented configuration; under the fixed-setting audit, this difference narrows to \textbf{0.25} points and is not significant after Holm correction. Meanwhile, random and unordered selection are \textbf{0.59} and \textbf{0.48} points below Full LAGO, respectively. These results provide behavioral support for confidence-aware guidance while more clearly supporting informative, ranked candidate selection.

\paragraph{Effect of object-context aggregation.}
Full LAGO exceeds the object-only and context-only variants by \textbf{0.53} and \textbf{0.54} points, respectively, supporting complementary local and contextual evidence. Object-focused crops preserve fine-grained details, while contextual regions and the full image retain broader layout information that tight crops may discard. This combination is particularly useful when recognition depends on both localized evidence and global structure.

\section{Conclusion}
We presented LAGO, a training-free framework that reframes localized zero-shot recognition as \emph{directed region discovery}. Rather than scoring a fixed set of regions after they are sampled, LAGO first establishes an object-centric visual initialization and then uses prediction-derived semantics to steer subsequent search, while preserving complementary object, context, and global evidence. Across standard and distribution-shift benchmarks, LAGO achieves strong zero-shot recognition performance and uses candidate regions more effectively, highlighting the value of language not only for interpreting visual evidence, but also for deciding where to look.

\paragraph{Limitations.}
LAGO's directed discovery depends on the quality of both the initial proposals and the prediction-derived semantic prototype; failures in either can steer refinement toward suboptimal regions. The two-stage search also increases cold-inference cost relative to WCA. Future work should investigate proposal-free discovery, more robust semantic prototypes, and leaner search strategies.


\subsection*{AI use statement}
In this work, we used generative AI tools to assist with drafting and editing parts of the manuscript, including language polishing, and with LaTeX formatting and layout. We did not use generative AI tools to generate or alter experimental data, design the methodology or experiments, implement the method, formulate mathematical claims or proofs, or interpret the experimental results. All AI-assisted text and formatting were reviewed and revised by the authors. We take responsibility for the final content of this work, including all text, claims, and artifacts produced with the aid of generative AI.

\subsection*{Ethics statement}
The human evaluation uses non-sensitive visual judgments over public benchmark images. Participation was voluntary; no personal information or behavioral data beyond multiple-choice annotation responses was collected. Under the applicable institutional policy, this minimal-risk internal annotation study did not require formal IRB or equivalent ethics-board review. Additional participant-facing details and the risk disclosure appear in Appendix~\ref{app:human_eval_instructions}.

\subsection*{Reproducibility statement}
The main paper specifies the model components and evaluation boundaries. Appendix~\ref{app:method_details} provides the exact proposal, search, confidence, text-construction, and aggregation settings; Appendix~\ref{app:adapter} separates source-domain specification, held-out calibration, full-test oracle diagnostics, and evaluation-label-calibrated results; Appendix~\ref{app:fixed_transfer} reports target-label-free transfer and the matched WCA comparison; and Appendices~\ref{app:ablation_details}--\ref{app:analysis} report component, robustness, statistical, runtime, and text-source audits.

\setlength{\parfillskip}{0pt plus 1fil}
{\small
\bibliographystyle{iclr2027_conference}
\bibliography{references}
}


\setlength{\parfillskip}{0pt plus 0.2\linewidth}
\appendix

\section{Additional Method Details}
\label{app:method_details}

\subsection{Offline Text Construction and Text Prototype Formation}
\label{app:text}

Following prior description-based zero-shot methods~\citep{menon2023visual, pratt2023platypus}, we maintain an offline class-specific description set for each class $y\in\mathcal{Y}$:
\begin{equation}
\mathcal{T}^{(y)}=
\{T_1^{(y)},T_2^{(y)},\dots,T_m^{(y)}\}.
\end{equation}
Each description encodes semantic properties such as parts, attributes, textures, or contextual cues and is embedded by the frozen CLIP text encoder $g(\cdot)$. These descriptions form a fixed text bank used for classification and Stage~2 semantic refinement.

We construct each class prototype as a weighted combination of its description embeddings. Let $\hat{T}_y$ denote the canonical prompt ``a photo of a \{class\}'', and let $\mathbf{u}_y=g(\hat{T}_y)$. The description weights are
\begin{equation}
a_i^{(y)}=
\frac{
\exp\!\left(\kappa\,\mathbf{u}_y^\top g(T_i^{(y)})\right)
}{
\sum_{j=1}^{m}
\exp\!\left(\kappa\,\mathbf{u}_y^\top g(T_j^{(y)})\right)
},
\end{equation}
and the class prototype is
\begin{equation}
\mathbf{t}_y=
\sum_{i=1}^{m}a_i^{(y)}g(T_i^{(y)}).
\end{equation}
Uniform weights $a_i^{(y)}=1/m$ recover standard mean pooling. In the reported CuPL experiments, we use the template-guided weighting above with $\kappa=e^2$. The resulting class prototypes form the classifier text bank $\mathbf{W}=[\mathbf{t}_y]_{y\in\mathcal{Y}}$.

After Stage~1, we construct a sample-specific semantic prototype from the intermediate logits $\mathbf{z}^{(1)}$. Let $\mathcal{K}$ index the five largest entries of $\mathbf{z}^{(1)}$. For $y\in\mathcal{K}$,
\begin{equation}
\pi_y=
\frac{
\exp(z_y^{(1)}/\tau_z)
}{
\sum_{y'\in\mathcal{K}}
\exp(z_{y'}^{(1)}/\tau_z)
},
\end{equation}
where $\tau_z$ controls the concentration over the leading class hypotheses. The Stage~2 prototype is
\begin{equation}
\mathbf{w}_{\mathrm{text}}
=
\sum_{y\in\mathcal{K}}\pi_y\mathbf{t}_y.
\end{equation}
This top-five mixture preserves uncertainty among the leading predictions and provides a smoother semantic target than committing refinement to a single predicted class.

\paragraph{Exact CuPL construction.}
We use the public CuPL class-level descriptions generated with GPT-3 \texttt{text-davinci-002}. For each class, CuPL uses five class-conditioned queries and samples ten completions per query with temperature $0.99$ and a maximum of 50 tokens. The queries are conditioned only on the class name; no test image, prediction, or image metadata is provided to the language model, and LAGO makes no per-image LLM call. We use the released descriptions without LAGO-specific manual editing.

\subsection{Detailed Proposal-Centered Search}
\label{app:search}

\paragraph{Proposal generation and filtering.}
FastSAM is used only as an off-the-shelf, class-agnostic proposal source. We use the released \texttt{FastSAM.pt} checkpoint without target-dataset training or adaptation, with inference size 1024, mask confidence threshold $0.4$, and FastSAM IoU threshold $0.9$. Masks are converted to axis-aligned boxes. We retain boxes covering between $5\%$ and $90\%$ of the image, apply score-ordered NMS at IoU $0.9$, and keep at most five proposal seeds. If no proposal survives, the full-image box is used. All boxes are clamped to the image boundary and cropped without padding, after which standard CLIP preprocessing is applied before feature extraction.

This section details the proposal-centered search routine used by the two-stage procedure in Sec.~\ref{sec:twostage}. Starting from each proposal seed, the routine performs coarse-to-fine local search over neighboring boxes using a stage-dependent scoring function.

For a candidate box $b$, we define the visual-consistency and semantic-alignment scores as
\begin{equation}
p(b)=\operatorname{sim}\!\left(\mathbf{v}(b),\mathbf{v}_{\mathrm{full}}\right),
\qquad
q(b)=\operatorname{sim}\!\left(\mathbf{v}(b),\mathbf{w}_{\mathrm{text}}\right),
\end{equation}
where all CLIP features are normalized. Stage~1 is purely visual and uses
\begin{equation}
S^{(1)}(b)=p(b),
\end{equation}
whereas Stage~2 uses
\begin{equation}
S^{(2)}(b)=(1-\gamma)p(b)+\gamma q(b),
\end{equation}
with the sample-specific guidance weight $\gamma$ determined from the intermediate confidence score. Thus, semantic information affects the search trajectory only after the class-agnostic Stage~1 initialization.

The top-ten Stage~1 crop features are aggregated to obtain $\mathbf{v}_{\mathrm{ens}}$, from which the intermediate class logits are computed as
\begin{equation}
\mathbf{z}^{(1)}=\mathbf{v}_{\mathrm{ens}}^{\top}\mathbf{W}.
\end{equation}
These logits define the prediction-derived semantic prototype $\mathbf{w}_{\mathrm{text}}$ through the top-five soft mixture described in Appendix~\ref{app:text}, preserving uncertainty among the leading class hypotheses rather than committing refinement to a single class.

Algorithm~\ref{alg:search_topk} summarizes the proposal-centered search routine shared by both stages.

\begin{algorithm}[t]
\caption{Proposal-Centered Coarse-to-Fine Crop Search}
\label{alg:search_topk}

\KwIn{
image $x$, initial box $b_0$, full-image feature $\mathbf{v}_{\mathrm{full}}$,
optional text prototype $\mathbf{w}_{\mathrm{text}}$, guidance weight $\gamma$
}
\KwOut{top-$K$ diverse crop set $\mathcal{C}$}

\Fn{\textnormal{\textsc{SearchTopKCrops}}
$(x,b_0,\mathbf{v}_{\mathrm{full}},\mathbf{w}_{\mathrm{text}},\gamma)$}{

    $b \gets \textsc{ClampToImage}(b_0)$\;
    $\mathcal{S} \gets \emptyset$\;

    compute $S_{\mathrm{visual}}(b)$\;
    \eIf{$\mathbf{w}_{\mathrm{text}}$ is available}{
        compute $S_{\mathrm{text}}(b)$\;
        $s_{\mathrm{cur}}
        \gets
        (1-\gamma)S_{\mathrm{visual}}(b)
        +\gamma S_{\mathrm{text}}(b)$\;
    }{
        $s_{\mathrm{cur}} \gets S_{\mathrm{visual}}(b)$\;
    }

    \ForEach{$\texttt{stage}\in\{\texttt{coarse},\texttt{fine}\}$}{

        $(T,\delta,\rho)
        \gets
        \textsc{StageParameters}(\texttt{stage})$\;

        \For{$t\gets1$ \KwTo $T$}{

            $\mathcal{N}
            \gets
            \textsc{GenerateNeighbors}(b,\delta,\rho)$\;

            \ForEach{$b'\in\mathcal{N}$}{

                $b' \gets \textsc{ClampToImage}(b')$\;
                compute $S_{\mathrm{visual}}(b')$\;

                \eIf{$\mathbf{w}_{\mathrm{text}}$ is available}{
                    compute $S_{\mathrm{text}}(b')$\;
                    $s(b')
                    \gets
                    (1-\gamma)S_{\mathrm{visual}}(b')
                    +\gamma S_{\mathrm{text}}(b')$\;
                }{
                    $s(b') \gets S_{\mathrm{visual}}(b')$\;
                }

                add $(b',s(b'))$ to $\mathcal{S}$\;
            }

            $\hat{b}
            \gets
            \arg\max_{b'\in\mathcal{N}} s(b')$\;

            \If{$s(\hat{b})-s_{\mathrm{cur}}<\epsilon$}{
                \KwBreak\;
            }

            $b \gets \hat{b}$\;
            $s_{\mathrm{cur}} \gets s(\hat{b})$\;
        }
    }

    $\mathcal{C}
    \gets
    \textsc{DiverseTopK}
    (\mathcal{S},K,\tau_{\mathrm{search}})$\;

    \KwRet{$\mathcal{C}$}\;
}
\end{algorithm}

\paragraph{Multi-start greedy refinement.}
We use multi-start greedy refinement initialized from the proposal set
$\mathcal{B}_0$. For each proposal seed $b_m^{(0)}$, the search iteratively
explores a local neighborhood and moves to the highest-scoring candidate:
\begin{equation}
b_{t+1}
=
\arg\max_{b' \in \mathcal{N}(b_t)} S(b').
\end{equation}
The update is performed independently for each seed for up to $T$ steps,
with early termination when the score improvement falls below $\epsilon$.

\paragraph{Neighborhood generation.}
For a current box
\begin{equation}
b=(x,y,w,h),
\end{equation}
local candidates are generated by translation and scale perturbations:
\begin{equation}
b'=
(x+\Delta x,\;
 y+\Delta y,\;
 w(1+\rho_w),\;
 h(1+\rho_h)).
\end{equation}
Here, $\Delta x$ and $\Delta y$ control spatial translation, while
$\rho_w$ and $\rho_h$ control scale variation. All generated boxes are
clamped to the image boundary before evaluation.

\paragraph{Coarse-to-fine search schedule.}
Each search trajectory consists of a coarse phase followed by a fine phase.
The coarse phase uses larger perturbations for broader exploration, whereas
the fine phase uses smaller perturbations for local refinement. In the
evaluated configuration, each phase runs for up to five steps, with
translation/scale step sizes of $0.15$ and $0.05$ for the coarse and fine
phases, respectively. Search terminates early when the score improvement is
below $10^{-4}$. At most eight candidate boxes are retained per seed within
each phase. After combining the two phases, exact integer-coordinate
duplicates are removed and candidates are processed with deterministic
IoU-based diversity filtering.

\paragraph{IoU-diverse top-$k$ selection.}
Candidates are sorted by score in descending order and greedily retained
only if their IoU with every previously selected box is below
$\tau_{\mathrm{search}}$. In the evaluated configuration,
$\tau_{\mathrm{search}}=0.5$, and at most 120 crop coordinates are retained
per image.

Each call to \textsc{SearchTopKCrops} therefore returns a diverse crop set
$\mathcal{C}$. Stage~1 applies the routine to proposal seeds in
$\mathcal{B}_0$ using the visual score alone to construct
$\mathcal{C}^{(1)}$. Stage~2 then applies the same routine to selected
Stage~1 regions using the prediction-derived semantic prototype, producing
$\mathcal{C}^{(2)}$ through language-guided refinement.

\paragraph{Complexity.}
Let $M$ denote the number of proposal seeds, $T$ the total number of search steps per seed, and $|\mathcal{N}|$ the neighborhood size. The geometric search evaluates
\begin{equation}
\mathcal{O}(MT|\mathcal{N}|)
\end{equation}
candidate boxes per stage, up to constant factors from the coarse-to-fine schedule. End-to-end cold inference additionally includes FastSAM proposal generation and CLIP encoding of the evaluated crops. For repeated evaluation on the same image, the resulting crop coordinates can be cached and reused without rerunning proposal generation and search.

\subsection{Additional View Diversity Control}
\label{app:view}

Crop-based inference can produce highly overlapping views that contribute redundant visual evidence. LAGO therefore applies IoU-based diversity control at two different points in the pipeline: within proposal-centered search and, when needed, during random completion to a fixed view budget.

\paragraph{IoU filtering for random completion crops.}
At inference time, the full image and proposal-guided crops are retained first. If fewer than the target number of views are available, additional random crops are sampled to fill the remaining slots. To avoid near-duplicate completion views, each sampled crop is accepted only if it is sufficiently distinct from all previously retained boxes.

Let $V$ denote the set of retained boxes, initialized with the full-image box and proposal-guided crops. A random candidate $b_i$ is accepted according to
\begin{equation}
f_{\mathrm{rand}}(b_i,V)=
\begin{cases}
1, & \mathrm{IoU}(b,b_i)<\tau_{\mathrm{rand}}
     \quad \forall b\in V,\\
0, & \text{otherwise}.
\end{cases}
\end{equation}
If $f_{\mathrm{rand}}(b_i,V)=1$, we update
$V\leftarrow V\cup\{b_i\}$; otherwise the candidate is discarded and another completion crop is sampled. This procedure preserves the full image and all proposal-guided views while increasing diversity among the additional context crops.

\paragraph{Relation to search-time diversity filtering.}
This completion filter is distinct from the IoU-diverse top-$k$ selection in Appendix~\ref{app:search}. Search-time filtering removes redundant candidates along proposal-centered trajectories before the Stage~1 or Stage~2 crop set is formed, whereas completion filtering acts only on additional random crops used to fill the remaining view budget. The two operations therefore serve the same diversity objective at different points in the inference pipeline.

\subsection{Fixed-Shape Crop Representation and Batched Inference}
\label{app:batch}

For batched inference, each image is represented by a fixed number of view
slots consisting of the full image, proposal-guided crops, and, when
necessary, random completion crops used to reach the target view budget.
If fewer valid views are available than the fixed tensor length, the
remaining slots are zero-padded and excluded from all subsequent
aggregation through a validity mask. This fixed-shape representation
enables straightforward batched evaluation without allowing padded views
to affect the final prediction.

\subsection{Confidence Mapping and Exact Aggregation}
\label{app:confidence_mapping}

\paragraph{Confidence mapping.}
Let $\mathcal{K}_{10}$ denote the indices of the ten largest intermediate
logits in $\mathbf{z}^{(1)}$. We normalize these logits as
\begin{equation}
\pi_j=
\frac{\exp(14.3\,z_j^{(1)})}
{\sum_{\ell\in\mathcal{K}_{10}}\exp(14.3\,z_\ell^{(1)})},
\qquad j\in\mathcal{K}_{10},
\end{equation}
and define the confidence score as
\begin{equation}
c=
\min\!\left(
1,\,
3\bigl(\pi_{(1)}-\pi_{(2)}\bigr)
\right),
\end{equation}
where $\pi_{(1)}$ and $\pi_{(2)}$ are the largest two normalized values.
The fixed scale $14.3\approx1/0.07$ controls the sharpness of the
intermediate distribution.

The semantic-guidance coefficient is then
\begin{equation}
\gamma(c)=
\gamma_{\min}
+
(\gamma_{\max}-\gamma_{\min})
\operatorname{clip}\!\left(
\frac{c-c_{\mathrm{low}}}
{c_{\mathrm{high}}-c_{\mathrm{low}}},
0,1
\right).
\end{equation}
The reference controller uses
$(c_{\mathrm{low}},c_{\mathrm{high}})=(0.08,0.30)$ and
$(\gamma_{\min},\gamma_{\max})=(0.20,0.80)$.
For the fixed-transfer protocol in Table~\ref{tab:fixed_wca_transfer},
we use the corresponding frozen range $0.20$--$0.65$ uniformly across
the evaluated targets. Appendix~\ref{app:mapping_ablation} evaluates
alternative mappings and shows limited sensitivity to the precise
functional form on the audited subset.

\paragraph{Exact aggregation.}
For completeness, the evaluated implementation aggregates the retained
crop features as
\begin{align}
r_i
&=
(1-\beta)p(c_i)+\beta q(c_i),
\\
w_i^{(\kappa)}
&=
\operatorname{softmax}_{i\in\mathcal{C}_{\kappa}}
\!\left(s_a r_i\right),
\\
\mathbf{v}_{\kappa}
&=
\operatorname{norm}\!\left(
\sum_{i\in\mathcal{C}_{\kappa}}
w_i^{(\kappa)}\mathbf{v}(c_i)
\right),
\qquad \kappa\in\{o,c\},
\\
\mathbf{v}_{\mathrm{dc}}
&=
\operatorname{norm}\!\left(
\alpha_{\mathrm{dc}}\mathbf{v}_{o}
+
(1-\alpha_{\mathrm{dc}})\mathbf{v}_{c}
\right),
\\
\mathbf{z}_{\mathrm{dc}}
&=
\mathbf{v}_{\mathrm{dc}}^{\top}\mathbf{W},
\\
\mathbf{z}_{\mathrm{final}}
&=
\lambda\mathbf{z}_{\mathrm{full}}
+
(1-\lambda)\mathbf{z}_{\mathrm{dc}}.
\end{align}
Here, $\mathcal{C}_{o}$ and $\mathcal{C}_{c}$ denote the object and context
crop sets, respectively; $\beta$ controls visual--semantic crop scoring,
$s_a$ controls the sharpness of crop weighting, and
$\alpha_{\mathrm{dc}}$ combines the two crop channels.
The coefficient $\lambda$ weights the full-image logits against the
dual-channel crop logits.

\subsection{Score-Level Calibration and Fixed-Transfer Protocol}
\label{app:adapter}

The fusion coefficients $(\beta,\alpha_{\mathrm{dc}},\lambda)$ can be
selected at the score level using frozen precomputed features, without
updating any model parameters. We distinguish two evaluation settings.
For the dataset-specific tuned results, the coefficients are selected
separately for each dataset--backbone configuration using evaluation
labels. These results therefore characterize tuned operating points rather
than target-label-free zero-shot transfer. Separately, we specify a single
coefficient tuple on ImageNet-V2 and freeze it before evaluation on all
transfer targets.

For the dataset-specific tuned protocol in
Table~\ref{tab:main_results}, we exhaustively evaluate 330 coefficient
tuples:
\begin{equation}
\beta\in\{0,0.1,\ldots,1.0\},\qquad
\alpha_{\mathrm{dc}}\in\{0,0.1,\ldots,0.5\},\qquad
\lambda\in\{0,0.01,0.025,0.05,0.1\}.
\end{equation}
Because this search operates only on precomputed scores, it requires no
encoder updates or additional feature extraction during the entire calibration procedure.

For the fixed-transfer protocol, ImageNet-V2 serves as the source
specification domain. The tuple
$(\beta,\alpha_{\mathrm{dc}},\lambda)=(1.0,0.30,0.025)$ is specified on
ImageNet-V2 and then frozen before evaluation on each of the eight transfer
targets in Appendix~\ref{app:fixed_transfer}; no target-domain labels are
used for coefficient selection. We additionally report held-out and
full-test oracle controls on ImageNet-V2 to quantify sensitivity to this
choice.

\begin{table}[!t]
\centering
\caption{\textbf{Protocol controls on ImageNet-V2.}
Held-out calibration and the source-specified tuple are evaluated on the
same disjoint 8,000-image report set for direct comparison.}
\label{tab:protocol_controls}
\small
\setlength{\tabcolsep}{4pt}
\begin{adjustbox}{max width=\linewidth,center}
\begin{tabular}{l l l c c}
\toprule
\textbf{Protocol}
& \textbf{Selection data}
& \textbf{Report set}
& $(\beta,\alpha_{\mathrm{dc}},\lambda)$
& \textbf{Acc.} \\
\midrule
Source-specified fixed tuple
& ImageNet-V2
& all 10,000
& $(1,.3,.025)$
& 65.45 \\

Held-out calibration
& 2,000 labels
& disjoint 8,000
& $(.7,0,.1)$
& 65.13 \\

Source-specified tuple
& prior ImageNet-V2 specification
& same 8,000
& $(1,.3,.025)$
& 65.11 \\

Full-test oracle
& all 10,000 labels
& all 10,000
& $(.9,.3,.05)$
& 65.69 \\
\bottomrule
\end{tabular}
\end{adjustbox}
\end{table}

On the shared 8,000-image report set, held-out calibration differs from
the source-specified tuple by only $0.02$ points. On the full ImageNet-V2
set, the oracle improves over the fixed tuple by $0.24$ points. These
controls indicate limited sensitivity to coefficient selection on the
source domain; cross-domain robustness is evaluated separately by freezing
the source-specified tuple before target evaluation.

\begin{table}[!t]
\centering
\caption{\textbf{One-at-a-time sensitivity on all 10,000 ImageNet-V2 images.}
The other two coefficients are held at the source-specified tuple
$(1.0,0.30,0.025)$ throughout the sensitivity analysis.}
\label{tab:parameter_sensitivity}
\small
\setlength{\tabcolsep}{5pt}
\begin{tabular}{l l c c}
\toprule
\textbf{Parameter}
& \textbf{Values}
& \textbf{Accuracy range}
& \textbf{Span} \\
\midrule
$\beta$
& $0,0.1,\ldots,1.0$
& 64.83--65.53
& 0.70 \\

$\alpha_{\mathrm{dc}}$
& $0,0.1,\ldots,0.5$
& 65.37--65.45
& 0.08 \\

$\lambda$
& $0,.01,.025,.05,.1$
& 65.40--65.49
& 0.09 \\
\bottomrule
\end{tabular}
\end{table}

The source-specified result is $65.45$, only $0.08$ points below the best
one-at-a-time setting. Across the tested ranges, sensitivity is modest for
$\alpha_{\mathrm{dc}}$ and $\lambda$, while $\beta$ has a larger effect.
Even at $\beta=0$, accuracy remains $64.83$, compared with $63.30$ for
full-image inference. Importantly, $\beta$ controls only final crop
aggregation; the Stage~2 guidance coefficient $\gamma$, which determines
language-guided region discovery, remains unchanged in this analysis.

\subsection{Cross-Dataset Transfer with a Fixed Tuple}
\label{app:fixed_transfer}

We evaluate whether a single source-specified operating point transfers
across datasets without target-domain coefficient selection. We specify
$(\beta,\alpha_{\mathrm{dc}},\lambda)=(1.0,0.30,0.025)$ on ImageNet-V2
and apply it unchanged to eight additional target datasets. No target-domain
labels are used to select or modify the fusion coefficients.

\begin{table}[t]
\centering
\caption{\textbf{Target-label-free transfer of a single ImageNet-V2-specified
fusion tuple.}
All rows use $(\beta,\alpha_{\mathrm{dc}},\lambda)=(1.0,0.30,0.025)$,
CuPL text, and no coefficient selection on the listed target.
The full-image comparator uses the same frozen CLIP backbone and
template-reweighted CuPL text bank as LAGO; it therefore differs from the
original CuPL baseline in Table~\ref{tab:main_results}.
$\Delta$ is relative to this matched full-image inference and is computed
before rounding. $\dagger$ denotes significance under exact McNemar testing
with Holm correction across the eight targets. Places365 uses the complete
36,500-image validation set with all 365 candidate classes.}
\label{tab:fixed_transfer}
\small
\setlength{\tabcolsep}{7pt}
\begin{adjustbox}{max width=\linewidth,center}
\begin{tabular}{l r c r r r}
\toprule
\textbf{Target dataset} & \textbf{$n$} & \textbf{Backbone}
& \textbf{Full image} & \textbf{Fixed LAGO}
& $\boldsymbol{\Delta}$ \\
\midrule
ImageNet
& 50,000 & ViT-L/14
& 76.62 & \textbf{77.63}$^\dagger$
& \textcolor{ForestGreen}{+1.01} \\

CUB
& 5,794 & ViT-L/14
& 63.95 & \textbf{64.48}
& \textcolor{ForestGreen}{+0.54} \\

DTD
& 1,880 & ViT-L/14
& 61.28 & \textbf{62.34}
& \textcolor{ForestGreen}{+1.06} \\

Food101
& 25,250 & ViT-L/14
& 93.53 & \textbf{93.66}
& \textcolor{ForestGreen}{+0.13} \\

Oxford Pets
& 3,669 & ViT-L/14
& 94.19 & \textbf{94.33}
& \textcolor{ForestGreen}{+0.14} \\

Places365
& 36,500 & ViT-B/32
& 43.54 & \textbf{44.32}$^\dagger$
& \textcolor{ForestGreen}{+0.78} \\

ImageNet-A
& 7,500 & ViT-B/16
& 50.93 & \textbf{57.95}$^\dagger$
& \textcolor{ForestGreen}{+7.01} \\

ImageNet-R
& 30,000 & ViT-B/16
& 77.30 & \textbf{77.50}
& \textcolor{ForestGreen}{+0.20} \\
\bottomrule
\end{tabular}
\end{adjustbox}
\end{table}

The source-specified tuple yields positive numerical differences over
matched full-image inference on all eight targets. The improvements on
ImageNet, Places365, and ImageNet-A remain significant after Holm
correction (largest adjusted $p=1.75\times10^{-4}$), while the remaining
five differences are positive but not significant after correction.
These results show that a single localized operating point can transfer
across diverse target domains without target-label coefficient selection.
Together with Table~\ref{tab:main_results}, the two protocols distinguish
dataset-specific tuned performance from fixed target-label-free transfer.

\paragraph{Matched-comparison view accounting.}
For the comparison in Table~\ref{tab:fixed_wca_transfer}, WCA evaluates
120 random views, whereas LAGO allocates 120 candidate-region slots and
encodes the full image separately, for up to 121 encoded views. The
comparison therefore matches the candidate-region budget rather than the
total number of encoded views. Random-completion crops fill unused
candidate slots. In the cached CUB, Places365, and ImageNet-A artifacts,
the mean number of valid candidate regions exceeds $119.96$; the minimum
is 103 on Places365, with invalid padded slots excluded by the validity
mask. The independently re-encoded DTD run uses 50 object and 70
random-context candidate slots with deterministic padding. The 48-view
setting in Appendix~\ref{app:runtime} and
Table~\ref{tab:efficiency_ablation} is a separate efficiency operating
point and is not used in this matched comparison.

\paragraph{Additional fixed-backbone audit.}
We further test whether the fixed-protocol trend is specific to a single
backbone by independently re-encoding DTD with CLIP ViT-B/16 and
ViT-L/14. Both runs use the same 1,880 images, candidate labels,
template-reweighted CuPL descriptions, and fixed tuple
$(1.0,0.30,0.025)$. WCA uses 120 deterministic random crops with seed 42,
whereas LAGO uses 50 object and 70 random-context candidate slots with
deterministic padding when fewer than 50 object crops are available.
No target labels are used for coefficient selection. Because these runs
are independently encoded, we report them separately from the cached
transfer results above.

\begin{table}[t]
\centering
\caption{\textbf{Fixed-tuple DTD comparison across two CLIP backbones.}
Differences and paired $95\%$ CIs are LAGO minus WCA; exact McNemar
$p$-values are Holm-corrected across the two backbone comparisons reported in this audit.}
\label{tab:dtd_backbone_transfer}
\small
\setlength{\tabcolsep}{4pt}
\begin{adjustbox}{max width=\linewidth,center}
\begin{tabular}{c r r r c c c}
\toprule
\textbf{Backbone}
& \textbf{Full image}
& \textbf{WCA@120}
& \textbf{LAGO@120}
& $\boldsymbol{\Delta}$ \textbf{[95\% CI]}
& \textbf{Exact $p$}
& \textbf{Holm $p$} \\
\midrule
ViT-B/16
& 51.17
& 52.77
& \textbf{53.51}
& \textcolor{ForestGreen}{+0.74 $[+0.00,+1.54]$}
& $0.0814$
& $0.163$ \\

ViT-L/14
& 61.28
& 61.70
& \textbf{62.34}
& \textcolor{ForestGreen}{+0.64 $[-0.21,+1.49]$}
& $0.182$
& $0.182$ \\
\bottomrule
\end{tabular}
\end{adjustbox}
\end{table}

Both backbones show positive numerical differences, although neither
comparison remains significant after Holm correction. The consistent
direction across the two independently encoded runs provides an additional
backbone-level robustness check for the fixed protocol.

\FloatBarrier
\begin{table}[t]
\centering
\caption{\textbf{Dataset- and backbone-specific fusion tuples used for the
dataset-specific tuned results in Table~\ref{tab:main_results}.}
Each tuple is selected from the score-level grid on the corresponding
evaluation cache and is reported for reproducibility.}
\label{tab:calibrated_tuples}
\small
\setlength{\tabcolsep}{7pt}
\begin{tabular}{lccc}
\toprule
\textbf{Dataset}
& \textbf{ViT-B/32}
& \textbf{ViT-B/16}
& \textbf{ViT-L/14} \\
\midrule
ImageNet
& $(1,.5,.1)$
& $(.9,.3,.1)$
& $(.8,.2,.1)$ \\

CUB
& $(.2,.1,.025)$
& $(.7,.3,0)$
& $(0,.4,0)$ \\

Oxford Pets
& $(0,.4,.1)$
& $(.9,.5,.025)$
& $(.4,.2,.01)$ \\

DTD
& $(1,.4,.1)$
& $(.4,.1,.05)$
& $(.9,0,.05)$ \\

Food101
& $(.3,.2,.1)$
& $(.3,.4,.1)$
& $(0,.1,.1)$ \\

Places365
& $(1,.3,.025)$
& $(.8,.3,.1)$
& $(1,.4,.1)$ \\
\bottomrule
\end{tabular}
\end{table}

\FloatBarrier
\section{Detailed Ablation Results}
\label{app:ablation_details}

Table~\ref{tab:ablation_studies_clean} gives the numerical values underlying the accuracy-oriented component summary in Figure~\ref{fig:ablation_effectiveness}. These values use evaluation-label score calibration and therefore quantify attainable component behavior rather than strict target-label-free improvements. Full LAGO reaches \textbf{65.90\%}; removing Stage~1 or Stage~2 reduces accuracy by $0.74$ and $0.32$ points, respectively, supporting the ordered visual-first, language-second design. Fixed guidance, random ranking, and unordered ranking reduce accuracy by $0.38$, $0.59$, and $0.48$ points. Object-only and context-only aggregation are $0.53$ and $0.54$ points below Full LAGO, indicating complementary object and contextual evidence within this calibrated audit.

Because calibration can magnify small component differences, Table~\ref{tab:fixed_component} repeats the central stage/guidance comparison under one frozen inference configuration. We treat the fixed audit as the primary evidence for the independent contribution of the confidence controller.

\begin{table}[H]
\centering
\captionsetup{skip=8pt}
\caption{\textbf{Evaluation-label-calibrated component audit on ImageNet-V2.}
Final, visual, object, and context accuracies are reported under the accuracy-oriented operating point. These post-hoc calibrated values are complementary to, and not interchangeable with, the frozen audit in Table~\ref{tab:fixed_component}.}
\label{tab:ablation_studies_clean}

\small
\setlength{\tabcolsep}{6pt}
\renewcommand{\arraystretch}{1.3}

\begin{adjustbox}{max width=\linewidth,center}
\begin{tabular}{l l l l c c c c}
\toprule
\textbf{Stage} & \textbf{Guidance} & \textbf{Ranking} & \textbf{Aggregation}
& \textbf{Final Acc (\%)} & \textbf{Visual Acc (\%)} & \textbf{Object Acc (\%)} & \textbf{Context Acc (\%)} \\
\midrule
Stage~1 only & -- & ranked & dual & 65.58 & 64.78 & 65.04 & 65.31 \\
Stage~2 only & adaptive & ranked & dual & 65.16 & 64.58 & 64.68 & 65.10 \\
Two-stage & fixed & ranked & dual & 65.52 & 64.63 & 65.12 & 65.28 \\
Two-stage & adaptive & random & dual & 65.31 & 64.55 & 64.87 & 65.10 \\
Two-stage & adaptive & unordered & dual & 65.42 & 64.55 & 65.29 & 65.19 \\
Two-stage & adaptive & ranked & object-only & 65.37 & 64.50 & 65.31 & 65.17 \\
Two-stage & adaptive & ranked & context-only & 65.36 & 64.76 & 65.25 & 65.28 \\
\rowcolor{gray!10}
Two-stage & adaptive & ranked & dual & \textbf{65.90} & \textbf{64.81} & \textbf{65.31} & \textbf{65.48} \\
\bottomrule
\end{tabular}
\end{adjustbox}
\end{table}

\FloatBarrier
\subsection{Fixed-Setting Component Simplification}
\label{app:fixed_component}

Table~\ref{tab:fixed_component} repeats the comparison under one fixed unified setting without target-label coefficient selection. Removing either stage reduces accuracy numerically by $0.25$--$0.32$ points, while replacing adaptive guidance with a constant weight reduces it by $0.25$ points. None of these paired contrasts is significant after Holm correction across the evaluated variants. The numerical pattern favors the complete ordered pipeline, but the fixed audit does not establish a statistically reliable independent contribution for any single component or for the particular confidence mapping.

\begin{table}[H]
\centering
\caption{\textbf{Frozen full-cache component audit on ImageNet-V2.} All variants share one inference configuration and use no target-label coefficient selection. $\Delta$ is relative to complete two-stage LAGO; none of the paired contrasts is significant after Holm correction.}
\label{tab:fixed_component}
\small
\setlength{\tabcolsep}{8pt}
\begin{tabular}{lcc}
\toprule
\textbf{Variant} & \textbf{Accuracy (\%)} & $\boldsymbol{\Delta}$ \\
\midrule
Complete two-stage LAGO & 65.45 & -- \\
Stage~1 only & 65.20 & $-0.25$ \\
Stage~2 only & 65.13 & $-0.32$ \\
Fixed guidance & 65.20 & $-0.25$ \\
\bottomrule
\end{tabular}
\end{table}

\FloatBarrier
\subsection{Directed-Discovery Mechanism Control}
\label{app:directed_discovery_control}

The controlled experiment in Table~\ref{tab:directed_discovery_control} uses a fixed 1,000-image CUB subset (seed 42), frozen OpenAI CLIP ViT-B/16, CuPL text, identical FastSAM proposals, identical Stage~1 candidates and ordering, identical Stage~2 seeds, and the fixed aggregation tuple $(\beta,\alpha_{\mathrm{dc}},\lambda)=(1.0,0.30,0.025)$. Early stopping is disabled, and every sample executes five coarse and five fine search steps. Post-hoc reranking sets $\gamma_{\mathrm{search}}=0$ and applies the shared semantic weight only during final selection; directed discovery uses the same weight during both search and selection. All audited shared inputs, Stage~1 outputs, intermediate logits, confidence values, proposal summaries, and final aggregation settings match exactly. Thus the intervention changes whether semantics steer the Stage~2 trajectory, rather than whether semantic reranking is available at the end.

\begin{table}[H]
\centering
\caption{\textbf{Controlled directed discovery versus post-hoc semantic reranking on 1,000 CUB images.} Both variants use frozen ViT-B/16 and share proposals, the $5{+}5$ search schedule, semantic final selection, 120 final slots, and aggregation. Differences use paired $95\%$ CIs; Query Top-3 is multiplied by 100 for readability.}
\label{tab:directed_discovery_control}
\small
\setlength{\tabcolsep}{5pt}
\begin{adjustbox}{max width=\linewidth,center}
\begin{tabular}{lccc}
\toprule
\textbf{Mechanism} & \textbf{Accuracy (\%)} & \textbf{Query Top-3 ($\times100$)} & \textbf{Changed crop sets} \\
\midrule
Post-hoc semantic reranking & 60.60 & 37.074 & -- \\
Directed semantic discovery & \textbf{61.40} & \textbf{37.117} & \textbf{786/1,000} \\
\midrule
Directed $-$ Post-hoc & $\mathbf{+0.80\;[+0.20,+1.50]}$ & $\mathbf{+0.043\;[+0.030,+0.057]}$ & -- \\
\bottomrule
\end{tabular}
\end{adjustbox}
\end{table}

The two trajectories have the same schedule, neighborhood operator, and 120 final candidate slots, but they do not evaluate an identical number of valid Stage~2 neighbors. Post-hoc reranking evaluates $644.579$ slots and $355.342$ unique boxes per image on average, compared with $652.096$ slots and $382.168$ unique boxes for directed discovery. The difference arises because the trajectories reach different locations, where boundary clipping and deduplication retain different numbers of valid neighbors. We therefore interpret this as a controlled search-policy comparison, not a strictly equal-compute comparison.

Directed discovery changes the final crop set on $786/1{,}000$ images; the mean and median set Jaccard similarities are $0.6465$ and $0.6062$. We measure query relevance with the text embedding of the shared Stage~1 top-1 prediction, which is fixed before the two Stage~2 variants diverge. Query Top-3 cosine increases from $0.370736$ to $0.371168$ (paired difference $+0.000432$, $95\%$ CI $[+0.000296,+0.000571]$), and query maximum cosine increases by $+0.000256$ ($[+0.000113,+0.000402]$). A label-aware diagnostic shows the same direction for ground-truth Top-3 ($+0.000418$, $[+0.000285,+0.000558]$) and maximum cosine ($+0.000227$, $[+0.000083,+0.000375]$), but ground-truth text is used only for analysis and never guides search.

For classification, directed discovery is correct on nine samples missed by post-hoc reranking, while the reverse occurs once. The accuracy difference is $+0.80$ points; 100,000 paired bootstrap replicates give $[+0.20,+1.50]$, and the two-sided exact McNemar value is $p=0.021484375$. Because this is the single prespecified mechanism comparison, no multiple-comparison correction is applied. The result establishes a reliable gain on this controlled subset, not on the complete CUB test set or across backbones and datasets. In an exploratory object-only aggregation that changes the prespecified final decision rule, post-hoc and directed variants obtain $59.10\%$ and $58.80\%$; we report this boundary case without using it to support the primary claim.

\FloatBarrier
\subsection{Proposal Robustness and Replacement}
\label{app:proposal_robustness}

We test whether LAGO remains useful when proposal coordinates are incomplete or noisy. Table~\ref{tab:proposal_corruption} reports an independently re-encoded, internally aligned ImageNet-V2 intervention audit. Its $65.27\%$ identity row is the reference for this table and should not be substituted for the $65.45\%$ canonical fixed-cache result. Relative to identity re-encoding, limiting the budget, deleting boxes, jittering coordinates, or adding distractors changes accuracy by at most $0.16$ points; none of these changes is significant after Holm correction across the evaluated perturbation variants.

\begin{table}[H]
\centering
\caption{\textbf{Coordinate-corruption audit on all 10,000 ImageNet-V2 images under matched re-encoding and controlled proposal perturbations applied to the same evaluation cache.}}
\label{tab:proposal_corruption}
\small
\setlength{\tabcolsep}{7pt}
\begin{tabular}{lc}
\toprule
\textbf{Coordinate condition} & \textbf{Accuracy (\%)} \\
\midrule
Identity re-encoding & 65.27 \\
Budget of 32 coordinates & 65.29 \\
Random deletion of $25\%$ & 65.17 \\
$10\%$ coordinate jitter & 65.21 \\
Add 20 distractors & 65.11 \\
\bottomrule
\end{tabular}
\end{table}

Complete replacement of FastSAM coordinates with generic grid or random boxes lowers absolute performance, but the downstream LAGO policy remains numerically above the matching proposal-only control (Table~\ref{tab:proposal_replacement}). Because this audit does not report paired uncertainty for the replacement contrasts, it shows that the behavior is not exclusive to FastSAM coordinates without establishing a statistically reliable gain for every proposal source.

\begin{table}[H]
\centering
\caption{\textbf{Complete proposal-replacement comparison with generic grid and random boxes on ImageNet-V2, using matched proposal-only controls for both coordinate sources.}}
\label{tab:proposal_replacement}
\small
\setlength{\tabcolsep}{7pt}
\begin{tabular}{lccc}
\toprule
\textbf{Coordinates} & \textbf{Proposal only} & \textbf{Full LAGO} & $\Delta$ \\
\midrule
Grid boxes & 63.39 & 64.56 & +1.17 \\
Random boxes & 63.68 & 64.58 & +0.90 \\
\bottomrule
\end{tabular}
\end{table}

On CUB, the FastSAM maximum-IoU mean/median is $0.700/0.718$, and its point-biserial correlation with classification correctness is $-0.0507$. On Oxford Pets, the corresponding values are $0.794/0.814$ and $-0.0432$. These near-zero correlations show little linear association between maximum proposal overlap and correctness in these audits; they do not imply that proposal coverage is irrelevant. If all candidates omit the decisive evidence, no downstream aggregation can recover it. Crowded, occluded, medical, satellite, or other specialized imagery may require denser multi-scale or domain-specific proposals.

\FloatBarrier
\subsection{Confidence-Mapping Alternatives and Behavioral Effects}
\label{app:mapping_ablation}

We compare four confidence-to-guidance mappings on an aligned 1,000-image ImageNet-V2 re-encoding (Table~\ref{tab:mapping_ablation}). Their point estimates are close on this subset, but the limited audit is not a general hyperparameter-robustness study. In particular, it does not establish a statistically reliable accuracy advantage of the linear rule over constant guidance.

\begin{table}[H]
\centering
\caption{\textbf{Alternative confidence mappings on an aligned 1,000-image ImageNet-V2 subset.}}
\label{tab:mapping_ablation}
\small
\setlength{\tabcolsep}{8pt}
\begin{tabular}{lc}
\toprule
\textbf{Inference rule} & \textbf{Accuracy (\%)} \\
\midrule
Full-image inference & 63.90 \\
Linear guidance & 66.60 \\
Sigmoid guidance & 66.60 \\
Threshold guidance & 66.50 \\
Constant guidance & 66.40 \\
\bottomrule
\end{tabular}
\end{table}

Linear guidance is $0.20$ points above constant guidance. For the reported contrast (linear minus constant), the paired $95\%$ confidence interval is $[-0.70,+1.10]$ and the exact McNemar value is $p=0.832$; no equivalence margin was prespecified, so non-significance is not evidence of equivalence. The two mappings produce 32 different predictions. Twenty-two changes affect correctness, all in the low-confidence group: linear guidance corrects 12 cases missed by constant guidance, while 10 change in the opposite direction. This supports a behavioral effect in the intended confidence regime, not a reliable aggregate-accuracy advantage.

\FloatBarrier
\section{Additional Analysis}
\label{app:analysis}

\subsection{Paired Statistical Evidence and Boundary Cases}
\label{app:statistical_evidence}

Fixed-cache inference is deterministic, so repeated nominal runs would have zero variance and would not quantify uncertainty. We instead align predictions by image path and label, use paired bootstrap confidence intervals for accuracy differences, and apply exact McNemar tests to paired correctness outcomes. For the four LAGO--WCA comparisons in Table~\ref{tab:fixed_wca_transfer}, we use 100,000 paired-image percentile bootstrap replicates with seed 0 and apply Holm correction to the four exact McNemar $p$-values. The additional DTD backbone audit applies the same procedure and corrects across its two rows. Table~\ref{tab:paired_statistics} retains complementary component comparisons and boundary cases.

\begin{table}[H]
\centering
\caption{\textbf{Path-aligned paired statistical analysis.} Differences are in percentage points. A dash indicates that no exact McNemar value was retained for that audit.}
\label{tab:paired_statistics}
\small
\setlength{\tabcolsep}{4pt}
\begin{adjustbox}{max width=\linewidth,center}
\begin{tabular}{l l r c c}
\toprule
\textbf{Dataset ($n$)} & \textbf{Reference} & $\Delta$ & \textbf{Paired 95\% CI} & \textbf{Exact McNemar} \\
\midrule
ImageNet-V2 (10,000) & visual initialization & +0.74 & $[+0.28,+1.19]$ & $p=0.00172$ \\
ImageNet-R (30,000) & visual initialization & -0.52 & $[-0.80,-0.23]$ & -- \\
Oxford Pets (3,669; ViT-L/14) & WCA & +0.055 & $[-0.164,+0.273]$ & $p=0.815$ \\
\bottomrule
\end{tabular}
\end{adjustbox}
\end{table}

The ImageNet-V2 interval excludes zero. In contrast, ImageNet-R is below visual initialization, and the Oxford Pets interval includes zero near the dataset's performance ceiling. The ImageNet-A matched result and its paired interval are reported once in Table~\ref{tab:fixed_wca_transfer}, avoiding a duplicate row with a separately generated bootstrap interval.

\FloatBarrier
\subsection{Stochastic Sources and Resampling}
\label{app:resampling}

To isolate actual stochasticity, we perform 20 within-cache completion resamples at each view budget. These draws come from a fixed candidate pool and do not rerun FastSAM or proposal-centered search. Table~\ref{tab:completion_resampling} shows that LAGO becomes more accurate than WCA at 40 views and above while exhibiting comparable or lower variability across resamples. This stochastic audit is distinct from the single matched budget sweep in Table~\ref{tab:efficiency_ablation}.

Prompt sampling is evaluated separately. With eight CuPL descriptions per class, five sampling seeds give $63.98\pm0.25$ for LAGO and $61.98\pm0.28$ for full-image inference. This variance reflects description sampling rather than model training or proposal generation.

\begin{table}[H]
\centering
\caption{\textbf{Within-cache completion resampling on ImageNet-V2.} We report mean top-1 accuracy and standard deviation across 20 resampled completions at each view budget.}
\label{tab:completion_resampling}
\small
\setlength{\tabcolsep}{8pt}
\begin{tabular}{ccc}
\toprule
\textbf{Views} & \textbf{WCA} & \textbf{LAGO} \\
\midrule
8  & $63.851\pm0.177$ & $62.400\pm0.000$ \\
16 & $64.214\pm0.176$ & $63.619\pm0.050$ \\
24 & $64.336\pm0.133$ & $63.930\pm0.078$ \\
32 & $64.453\pm0.130$ & $64.329\pm0.079$ \\
40 & $64.465\pm0.103$ & $64.543\pm0.084$ \\
48 & $64.489\pm0.080$ & $64.728\pm0.071$ \\
64 & $64.513\pm0.104$ & $64.964\pm0.065$ \\
\bottomrule
\end{tabular}
\end{table}

\FloatBarrier
\subsection{End-to-End Runtime and Deployment Cost}
\label{app:runtime}

We measure two non-overlapping runtime boundaries on one A100-SXM4-80GB GPU in FP32 with batch size 1, five warm-up images, and 50 timed images. Every per-image timer starts from compressed-image bytes and includes image decoding and preprocessing.

\begin{table}[H]
\centering
\caption{\textbf{Cold per-image construction time.} LAGO includes FastSAM and both search stages.}
\label{tab:cold_runtime}
\small
\setlength{\tabcolsep}{7pt}
\begin{tabular}{lc}
\toprule
\textbf{Method} & \textbf{Latency (ms/image)} \\
\midrule
Global CLIP & 14.46 \\
Waffle & 14.31 \\
WCA@120 & 404.37 \\
LAGO cold construction & 1702.02 \\
\bottomrule
\end{tabular}
\end{table}

The LAGO cold total comprises FastSAM ($107.48$ ms), proposal ranking ($16.71$ ms), two-stage search ($1107.13$ ms), final crop encoding ($451.75$ ms), full-image encoding ($14.96$ ms), and decoding/aggregation ($3.99$ ms). Thus cold LAGO is $4.21\times$ slower than WCA@120 ($1702.02$ versus $404.37$ ms/image) and must not be described as generally runtime-efficient. CuPL text encoding takes $18.665$ s once per dataset; CLIP and FastSAM loading take $1.696$ s and $0.910$ s once per process.

\begin{table}[H]
\centering
\caption{\textbf{Accuracy-matched repeated-query deployment.} Ranked coordinates are reusable, but all image decoding and encoding remain online for every repeated prediction. This is a separate deployment operating point, not the deterministic budget-sweep row in Table~\ref{tab:efficiency_ablation}.}
\label{tab:deployment_runtime}
\small
\setlength{\tabcolsep}{6pt}
\begin{tabular}{lcccc}
\toprule
\textbf{Method} & \textbf{Views} & \textbf{Accuracy} & \textbf{Latency} & \textbf{Throughput} \\
 & & (\%) & (ms/image) & (image/s) \\
\midrule
WCA & 120 & 64.52 & 390.99 & 2.56 \\
LAGO & 48 & 64.73 & 189.00 & 5.29 \\
\bottomrule
\end{tabular}
\end{table}

At this repeated-query operating point, LAGO uses $60\%$ fewer views, reduces online latency by $51.66\%$, and is $2.07\times$ faster after ranked coordinates have been constructed. The reusable state contains coordinates only, never cached features or predictions. This amortized scenario does not remove the $4.21\times$ larger one-time cold construction cost and is not presented as standard one-pass end-to-end efficiency for ordinary classification workloads.

\FloatBarrier
\subsection{Text-Description Dependence}
\label{app:text_dependence}

We replace CuPL with four alternative text sources while keeping the localized inference mechanism unchanged. Table~\ref{tab:text_sources} shows that description richness affects absolute accuracy; every tested source has a positive numerical localized difference, including class names without LLM-generated descriptions. We do not attach a separate significance claim to these five text-source contrasts.

\begin{table}[H]
\centering
\caption{\textbf{Dependence of ImageNet-V2 recognition accuracy on the class-text source.}}
\label{tab:text_sources}
\small
\setlength{\tabcolsep}{8pt}
\begin{tabular}{lccc}
\toprule
\textbf{Text source} & \textbf{Full image} & \textbf{LAGO} & \textbf{Gain} \\
\midrule
Class name & 57.81 & 60.35 & +2.54 \\
Single CLIP template & 60.72 & 62.92 & +2.20 \\
CLIP-E & 61.99 & 64.46 & +2.47 \\
CLIP-D & 61.50 & 63.37 & +1.87 \\
CuPL & 63.30 & 65.45 & +2.15 \\
\bottomrule
\end{tabular}
\end{table}

The CuPL row above uses the full released description bank and therefore matches the canonical fixed-cache result of $65.45\%$. The following count study deliberately subsamples that bank. For a fixed sampling seed, using $1/2/4/8/16$ CuPL descriptions per class gives LAGO accuracies of $53.78/59.27/63.08/64.39/65.05$, respectively. At eight descriptions, five seeds give $63.98\pm0.25$ (range $63.74$--$64.39$); the corresponding full-image result is $61.98\pm0.28$. Thus the absolute level depends on description coverage, while the localized improvement is not an artifact of a single prompt source under this controlled comparison.

\FloatBarrier
\subsection{Prediction-Conditioned Feedback Audit}
\label{app:feedback_statistics}

Stage~2 constructs semantic guidance once from the Stage~1 logits, so an inaccurate intermediate hypothesis can influence the regions used by the final predictor. To match Algorithm~\ref{alg:two_stage}, let $R^{(1)}$ denote the visual-only regions and write
\begin{align}
\mathbf{z}^{(1)} &= \Psi(x[R^{(1)}],\mathbf{W}),
&\hat{y}^{(1)} &= \arg\max_y z^{(1)}_y,\\
\mathbf{w}_{\mathrm{text}} &= G(\mathbf{z}^{(1)},\mathbf{W}),
&\gamma &= \psi\!\left(\phi(\mathbf{z}^{(1)})\right),\\
R^{(2)} &= \Phi(R^{(1)},\mathbf{w}_{\mathrm{text}},\gamma),
&\mathbf{z}^{(f)} &= \Omega(x,R^{(1)},R^{(2)},\mathbf{W}).
\end{align}
Here $G$ is the soft-prototype construction in Appendix~\ref{app:text}, $\phi$ and $\psi$ form the confidence controller, $\Phi$ is semantic region refinement, and $\Omega$ is the final object-context/full-image aggregation. The prototype and guidance weight remain fixed during the Stage~2 local search; they are not recomputed after every neighborhood update.

For an initially wrong sample, persistence records whether the final prediction retains the same wrong class:
\begin{equation}
\mathbb{1}[\hat{y}^{(f)}=\hat{y}^{(1)}],
\qquad
\hat{y}^{(f)}=\arg\max_y z^{(f)}_y.
\end{equation}
Amplification records whether the final predictor increases the initial wrong-class margin relative to the ground-truth class:
\begin{equation}
\mathbb{1}\!\left[
(z^{(f)}_{\hat{y}^{(1)}}-z^{(f)}_{y^*})
-(z^{(1)}_{\hat{y}^{(1)}}-z^{(1)}_{y^*})>0
\right].
\end{equation}
This is a two-stage prediction-conditioned dependence, not an online recurrent prediction loop within every search step.

On all 10,000 ImageNet-V2 images, visual initialization makes 3,549 errors. Among these errors, 2,829 ($79.71\%$) retain the same wrong class after the complete refinement and aggregation pipeline, and 1,326 ($37.36\%$) amplify the initial wrong-versus-true margin. LAGO corrects 309 initially wrong samples and induces 235 errors among the 6,451 initially correct samples, yielding 74 net additional correct predictions. These end-to-end transition statistics show that prediction-conditioned refinement can both correct and preserve an initial error. They motivate visual initialization and cautious semantic guidance, but do not isolate directed search, aggregation, or the confidence controller; the controlled comparison in Appendix~\ref{app:directed_discovery_control} addresses the search-policy effect separately.

\FloatBarrier
\subsection{Comparison with Recent Localized Systems}
\label{app:recent_baselines}

We evaluate cited localized methods with public runnable releases on the full 10,000-image ImageNet-V2 set. Because FG-CLIP~\citep{xie2025fgclip} and FLAIR~\citep{xiao2025flair} use learned encoders and additional pretraining, Table~\ref{tab:recent_localized} is a released-system comparison rather than a controlled same-encoder ablation. The controlled claim remains the improvement over localized inference methods using the same frozen OpenAI CLIP backbone for careful attribution.

\begin{table}[H]
\centering
\caption{\textbf{Released-system comparison on the full ImageNet-V2 evaluation set under official protocols, using each method's public runnable release and its stated evaluation configuration.}}
\label{tab:recent_localized}
\small
\setlength{\tabcolsep}{5pt}
\begin{adjustbox}{max width=\linewidth,center}
\begin{tabular}{l l c}
\toprule
\textbf{Method} & \textbf{Protocol} & \textbf{Accuracy (\%)} \\
\midrule
LAGO & frozen OpenAI ViT-B/16, CuPL, inference only & 65.45 \\
FG-CLIP official & base, one template & 61.32 \\
FG-CLIP official & base, 80 templates & 61.74 \\
FLAIR official & ViT-B/16-FLAIR, top 256 & 49.01 \\
FLAIR official & ViT-B/16-FLAIR, global & 45.95 \\
\bottomrule
\end{tabular}
\end{adjustbox}
\end{table}

Under these released configurations, LAGO is $3.51$ points above the strongest FG-CLIP setting and $16.24$ points above FLAIR top-256. Because encoders, pretraining, prompts, and official protocols differ, these numbers compare runnable systems rather than attribute the differences to LAGO's inference policy. We omit localized systems without public code or checkpoints rather than report irreproducible numerical comparisons~\citep{chen2025cloc}.

\FloatBarrier
\subsection{Candidate-Region Efficiency}
\label{app:efficiency}

\begin{table}[H]
\centering
\captionsetup{skip=6pt}
\caption{\textbf{Candidate-region efficiency on ImageNet-V2.} We report one matched budget sweep with the same protocol for WCA and LAGO at every listed budget. Absolute values from the separate completion-resampling and deployment audits should not be substituted into this table.}
\label{tab:efficiency_ablation}
\small
\setlength{\tabcolsep}{6pt}
\renewcommand{\arraystretch}{1.15}
\begin{tabular}{lcccccc}
\toprule
\textbf{Method} & \textbf{8} & \textbf{16} & \textbf{24} & \textbf{32} & \textbf{48} & \textbf{64} \\
\midrule
WCA  & 63.96 & 64.44 & 64.44 & 64.53 & 64.55 & 64.44 \\
LAGO & 63.48 & \textbf{64.49} & \textbf{64.88} & \textbf{65.06} & \textbf{65.33} & \textbf{65.56} \\
\midrule
$\Delta$ & -0.48 & \textcolor{ForestGreen}{+0.05} & \textcolor{ForestGreen}{+0.44} & \textcolor{ForestGreen}{+0.53} & \textcolor{ForestGreen}{+0.78} & \textcolor{ForestGreen}{+1.12} \\
\bottomrule
\end{tabular}
\end{table}

We further analyze whether LAGO uses a fixed candidate-region budget more effectively than WCA, where the budget denotes the total number of candidate regions evaluated per image at inference time. Table~\ref{tab:efficiency_ablation} shows that LAGO is slightly weaker at the smallest budget because too few regions leave limited room for proposal-centered search to establish sufficient coverage before refinement. LAGO becomes stronger as soon as the budget reaches 16 regions, and the gap then widens through 64 regions. This pattern supports increasingly informative and selective region discovery under a fixed inference budget, while also exposing the minimum coverage needed by the proposal-centered search.

\subsection{Confidence-Aware Behavior}
\label{app:confidence_behavior}

We provide a behavioral analysis of confidence-aware refinement. To identify where adaptive and fixed guidance differ, we group samples by confidence and examine cases corrected by the adaptive variant but not by the fixed-guidance baseline. This conditional subset does not measure the controller's overall accuracy effect or its statistical significance.

Figure~\ref{fig:confidence_failure_analysis} (left) summarizes adaptive-only corrections across confidence buckets. Most occur in the \emph{low-confidence} regime, with fewer in the \emph{medium-confidence} regime. This concentration is consistent with the intended behavior: uncertain intermediate predictions receive more cautious semantic guidance. It is not evidence of a statistically reliable aggregate advantage, because the reverse changes and the fixed-setting significance tests must also be considered.

Figure~\ref{fig:confidence_case_grid} presents adaptive-only successes and therefore provides illustrative, one-sided evidence. The separate linear-versus-constant audit in Appendix~\ref{app:mapping_ablation} reports changes in both directions (12 corrections versus 10 regressions). Together, the results show that confidence changes decisions in its intended regime without establishing a significant aggregate-accuracy improvement.

\subsection{Stage-wise Visualization}
\label{app:stage_vis}

We provide qualitative examples of how region discovery evolves from Stage~1 to Stage~2. Stage~1 performs visual-only localization and tends to select salient object-centric crops without assuming reliable class semantics. Stage~2 then refines these regions with text guidance, often shifting attention toward evidence associated with the predicted class. These examples visualize the ordered design---first stabilize localization, then refine it semantically---but do not isolate the controller's aggregate accuracy contribution under fixed inference.

\begin{figure}[!t]
    \centering
    \includegraphics[width=\textwidth]{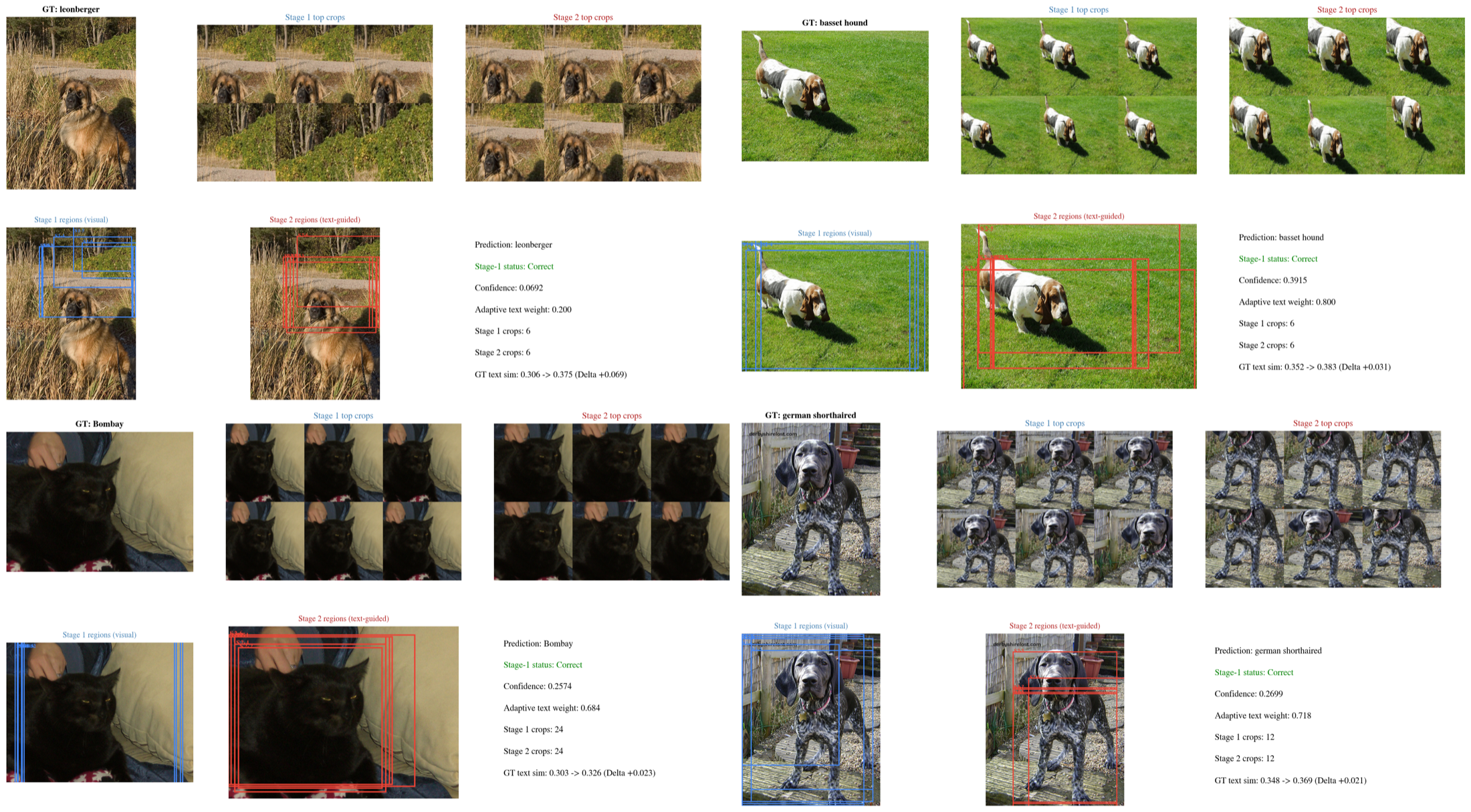}
    \caption{\textbf{Stage-wise visualization on Oxford Pets.}
    Each example shows the image, top crops from Stage~1 and Stage~2, their candidate regions, and statistics for confidence and text-similarity changes. Stage~1 selects object-centric regions, while Stage~2 uses the prediction-derived text prototype to move the search toward more semantically informative, class-relevant evidence.}
    \label{fig:stage_vis_pets}
\end{figure}

Figure~\ref{fig:stage_vis_pets} highlights the refinement process with detailed per-example statistics, while Figure~\ref{fig:stage_vis_multi} illustrates the same qualitative pattern on selected examples from multiple datasets. These figures show behavior, not cross-dataset effect sizes.

\subsection{Failure Analysis: ImageNet-R Case Study}
\label{app:failure}

We provide a qualitative failure analysis using ImageNet-R as a case study for zero-shot localized alignment analysis. We choose ImageNet-R because its artistic and stylized renditions, where appearance cues are often semantically misleading, make the alignment between class semantics and visually relevant regions particularly challenging under natural distribution shift at inference time.

\begin{figure}[!t]
    \centering
    \begin{subfigure}{0.55\linewidth}
        \centering
        \includegraphics[width=\linewidth]{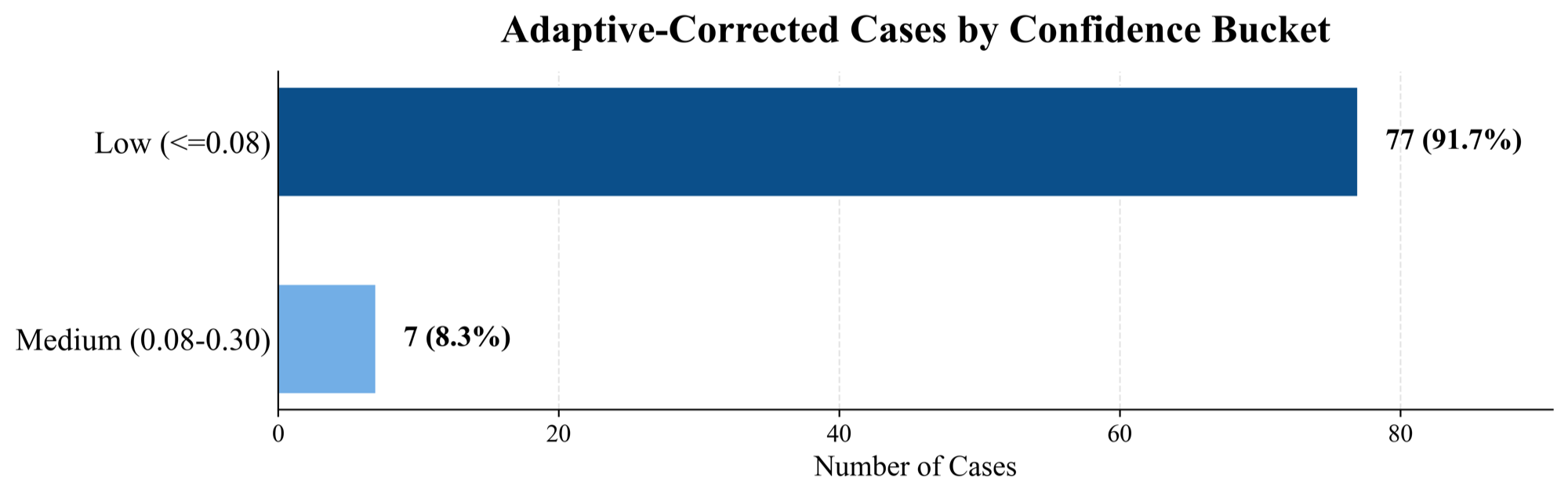}
        \caption{Adaptive-corrected cases by confidence bucket}
        \label{fig:confidence_bucket_bar}
    \end{subfigure}
    \hfill
    \begin{subfigure}{0.44\linewidth}
        \centering
        \includegraphics[width=\linewidth]{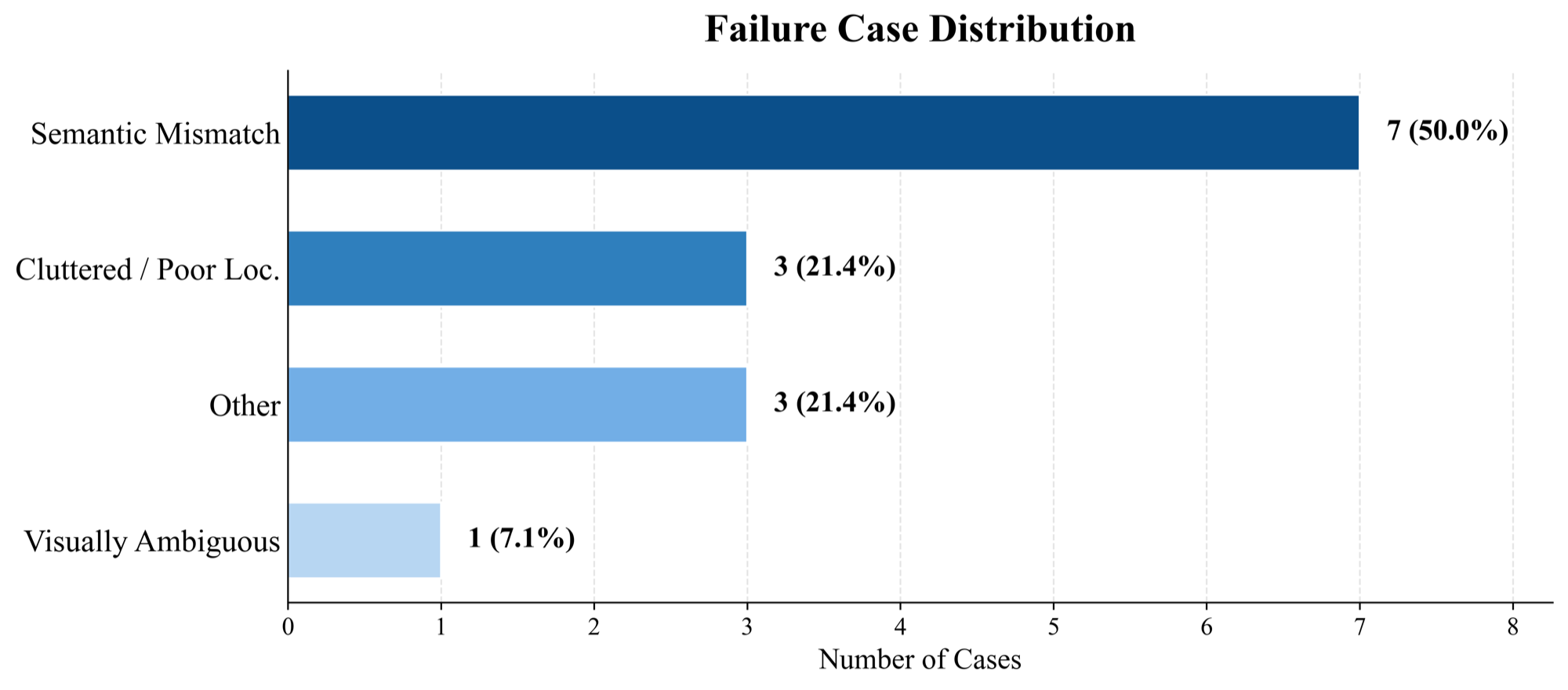}
        \caption{Categories in the inspected ImageNet-R failure set}
        \label{fig:failure_dist}
    \end{subfigure}

    \caption{\textbf{Confidence behavior and qualitative failure audit.} \textbf{Left:} Adaptive-only corrections across confidence buckets. \textbf{Right:} Category proportions within the inspected ImageNet-R failure set; these are not estimates of dataset-wide prevalence.}
    \label{fig:confidence_failure_analysis}
\end{figure}

\begin{figure}[!t]
    \centering
    \includegraphics[width=\textwidth]{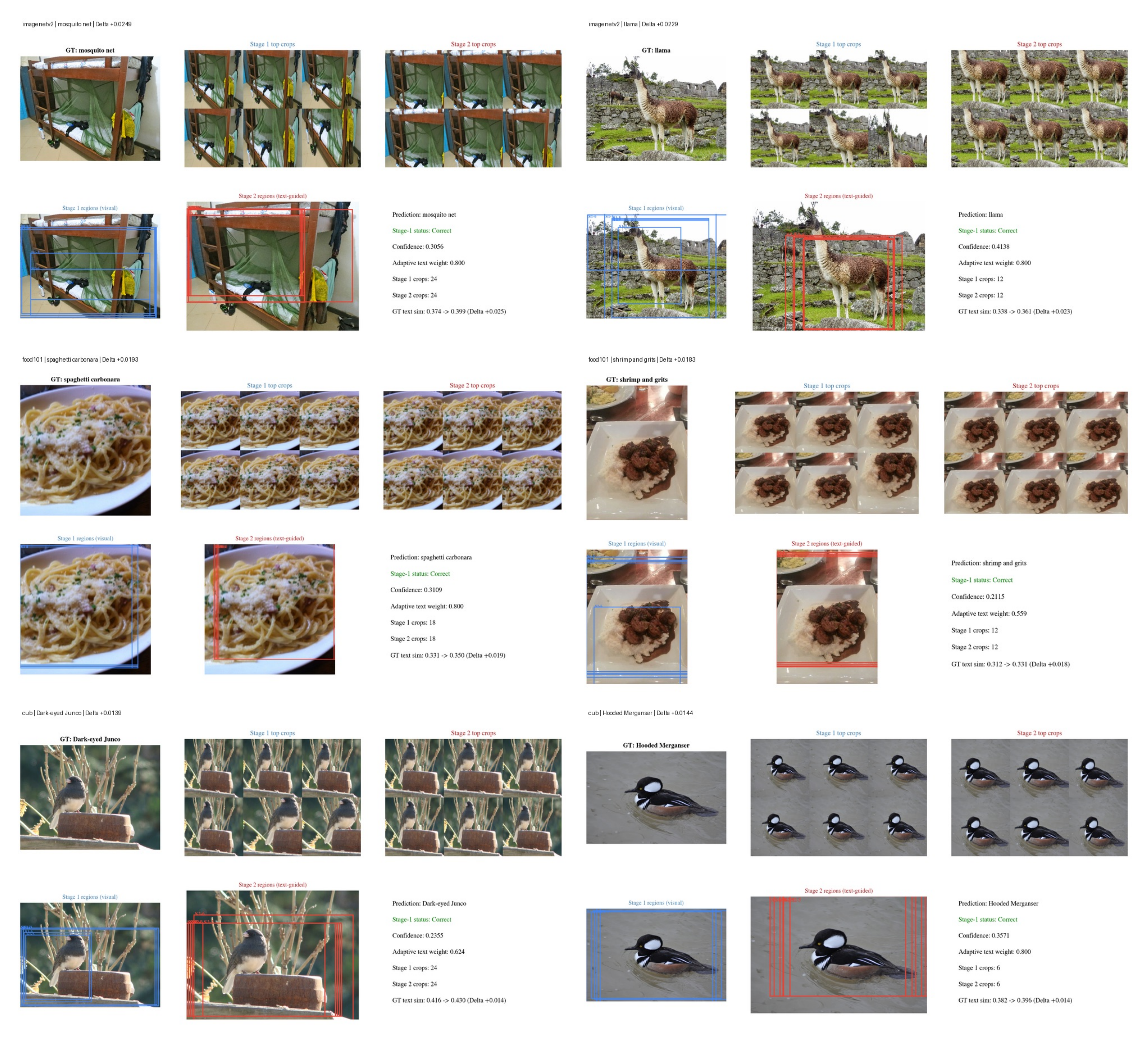}
    \caption{\textbf{Stage-wise visualization across ImageNet-V2, Food101, and CUB.}
    Representative examples from multiple datasets show a consistent pattern: Stage~1 identifies visually salient and object-centric regions, whereas Stage~2 shifts the selected crops toward evidence associated with the predicted class. This figure illustrates the ordered refinement behavior across generic object recognition, food classification, and fine-grained bird recognition; it does not isolate the aggregate accuracy contribution of confidence modulation.}
    \label{fig:stage_vis_multi}
\end{figure}

\begin{figure*}[t]
    \centering
    \includegraphics[width=\textwidth]{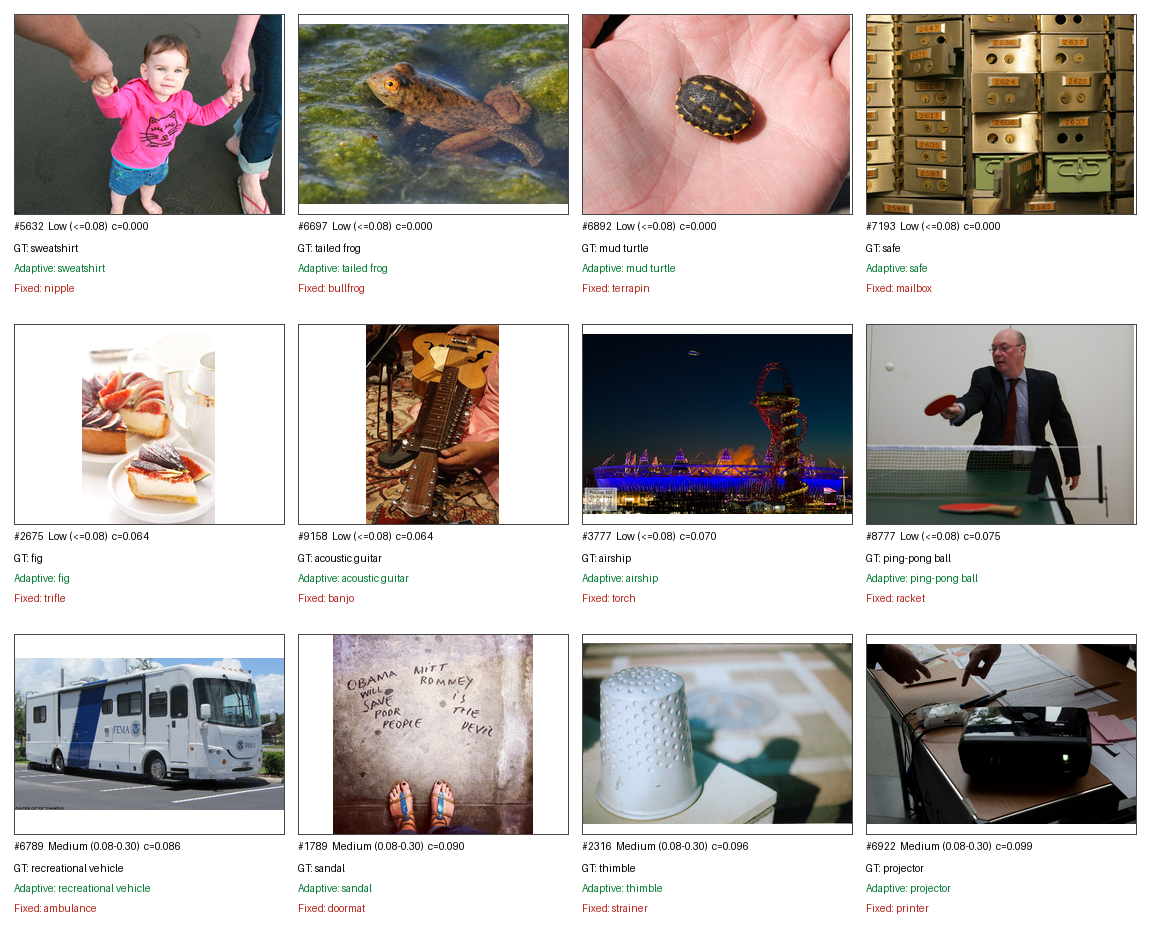}
    \caption{\textbf{Representative cases corrected by adaptive guidance.}
    Each example shows an image for which the adaptive strategy predicts the correct class while the fixed-guidance variant fails at inference time. These adaptive-only successes are grouped by intermediate-confidence bucket and are concentrated in the low-confidence regime. Because reverse changes are not shown here, the examples provide behavioral evidence rather than an aggregate-accuracy comparison.}
    \label{fig:confidence_case_grid}
\end{figure*}

\begin{figure*}[t]
    \centering
    \includegraphics[width=\textwidth]{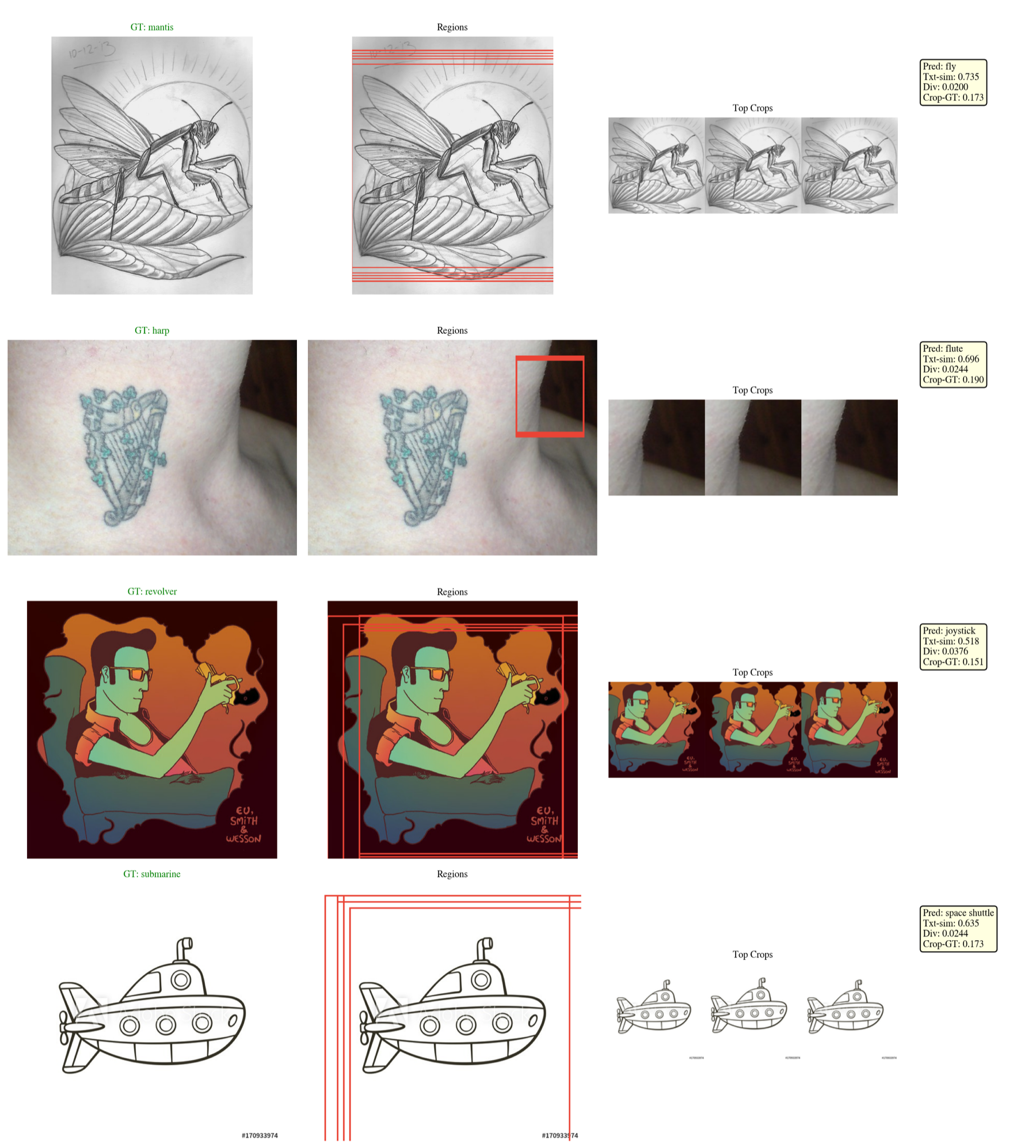}
    \caption{\textbf{Representative semantic mismatch failures on ImageNet-R.} The selected regions are locally plausible, but the text-side semantics emphasize misleading cues from the predicted text prototype under distribution shift conditions that do not support the ground-truth category.}
    \label{fig:failure_semantic}
\end{figure*}

\begin{figure*}[t]
    \centering
    \includegraphics[width=\textwidth]{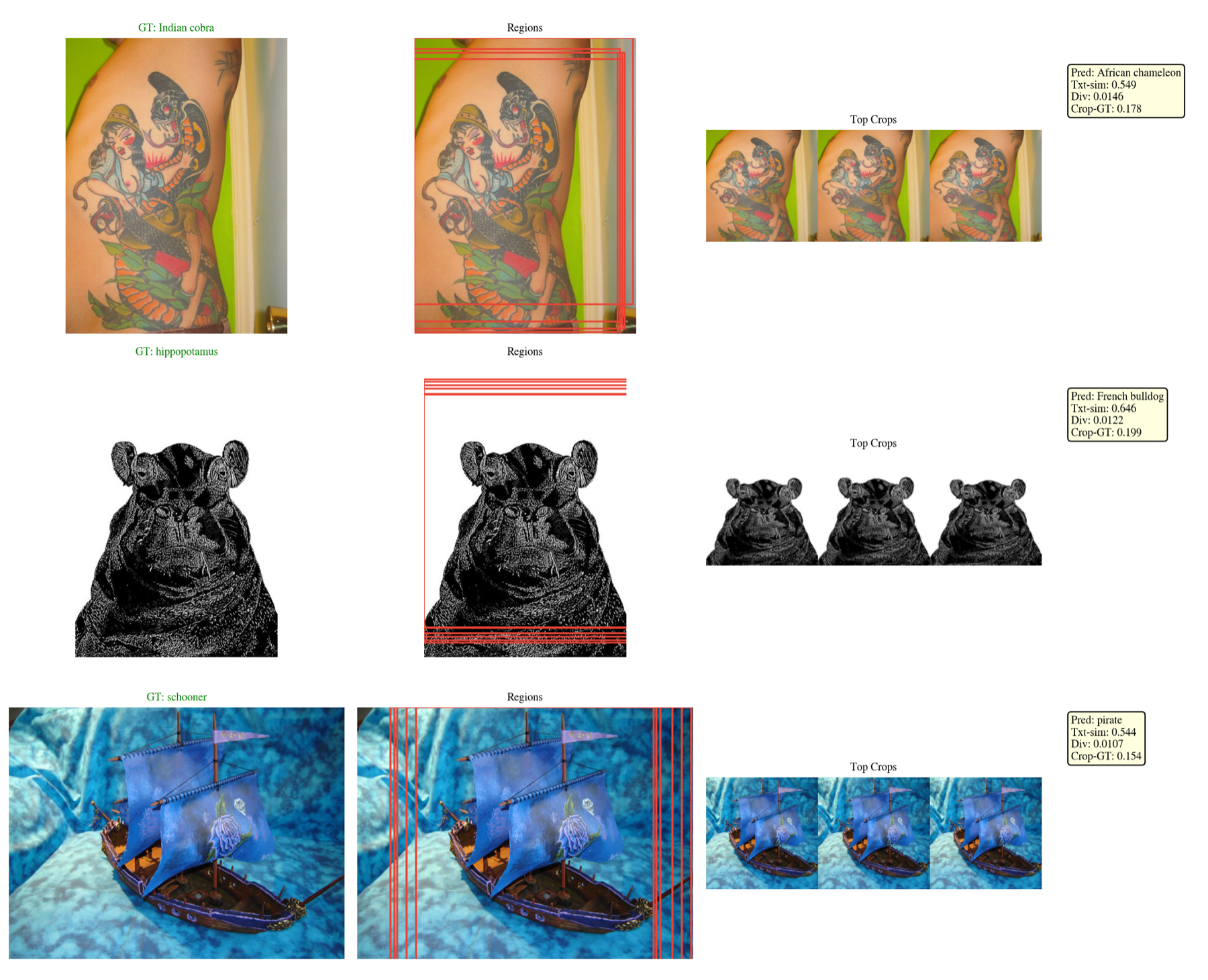}
    \caption{\textbf{Representative cluttered or poorly localized failures on ImageNet-R.} In these selected examples, the proposal set omits or incompletely captures discriminative evidence, leaving later semantic refinement with weak or irrelevant regions.}
    \label{fig:failure_cluttered}
\end{figure*}

\begin{figure*}[t]
    \centering
    \includegraphics[width=\textwidth]{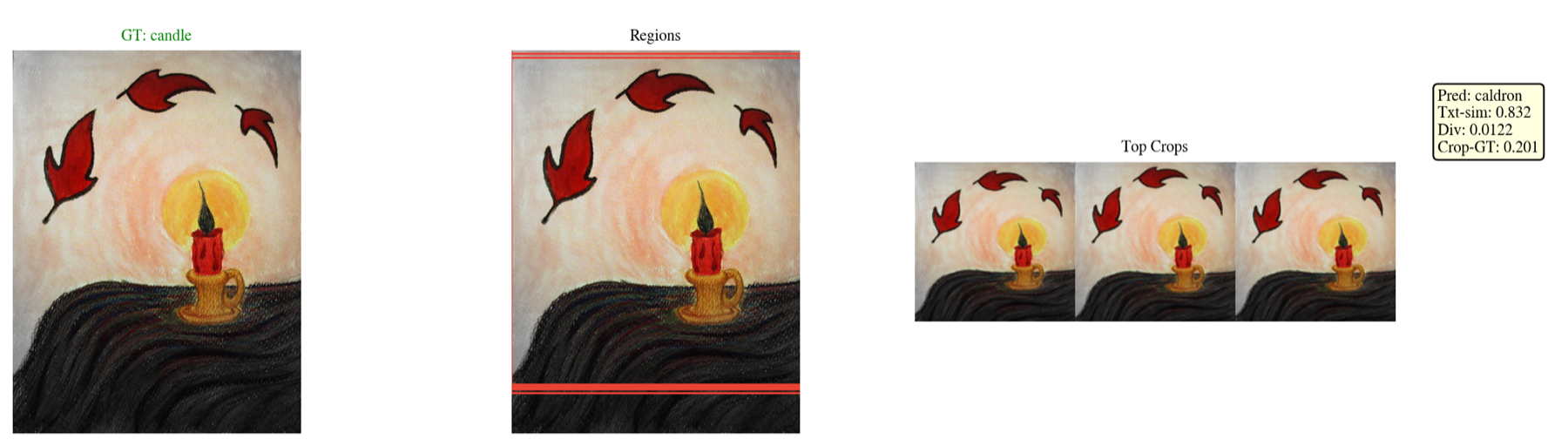}
    \caption{\textbf{Representative visually ambiguous failures on ImageNet-R.} Even after localized refinement, visually similar categories remain difficult to distinguish because they share highly overlapping local and semantically similar attributes under natural distribution shift at inference time.}
    \label{fig:failure_ambiguous}
\end{figure*}

\begin{figure*}[t]
    \centering
    \includegraphics[width=\textwidth]{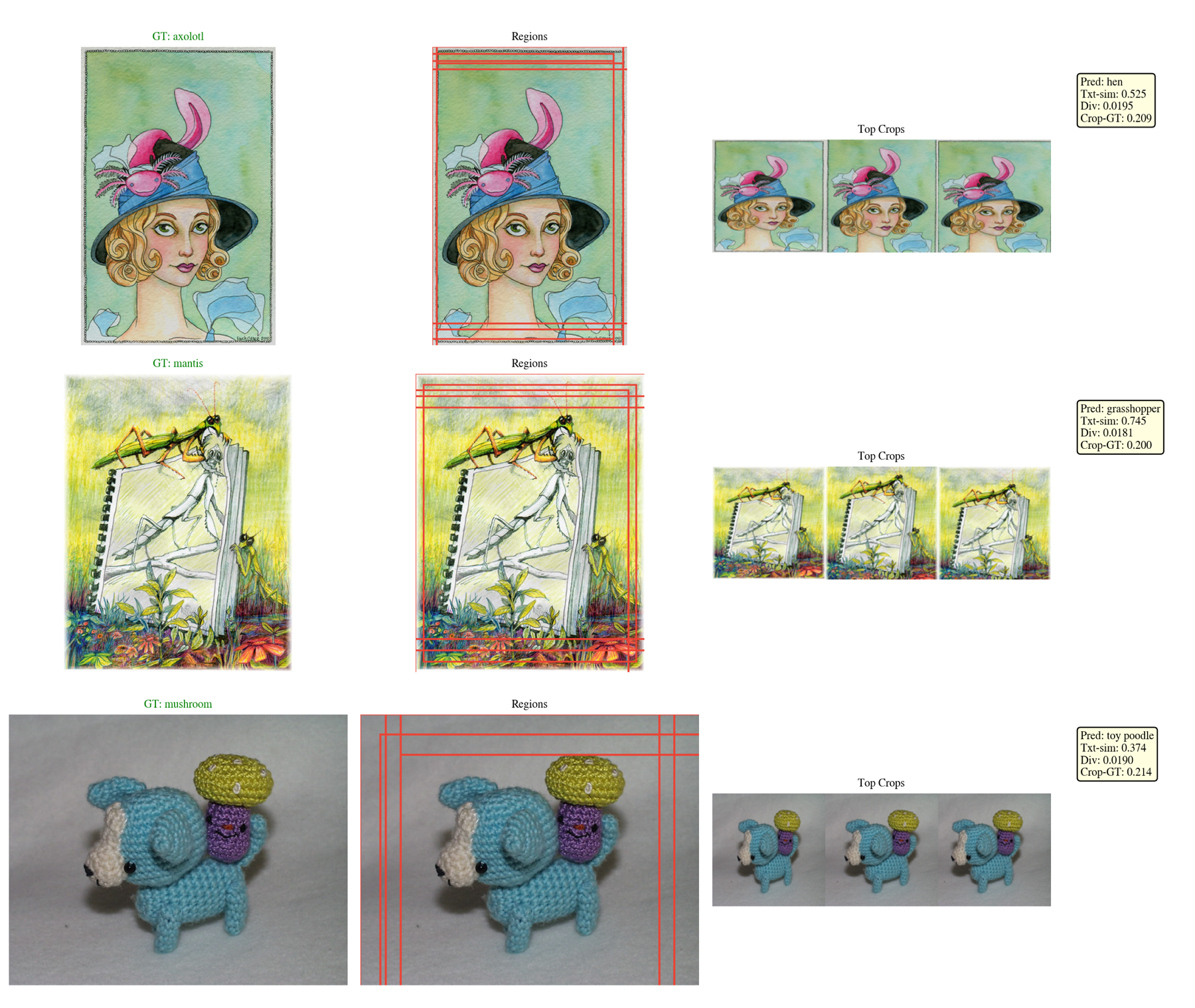}
    \caption{\textbf{Representative failure cases categorized as other on ImageNet-R.} These examples do not fall cleanly into a single dominant error type, but still illustrate challenging cases for localized visual-text alignment under natural distribution shift with mixed semantic and localization errors.}
    \label{fig:failure_other}
\end{figure*}

Figure~\ref{fig:confidence_failure_analysis} (right) summarizes categories within the inspected ImageNet-R failure set. Semantic mismatch is the largest category in this qualitative sample, followed by cluttered or poorly localized regions, other cases, and visually ambiguous classes. Figures~\ref{fig:failure_semantic}--\ref{fig:failure_other} show representative examples. The proportions characterize the inspected cases and are not estimates of dataset-wide failure prevalence.

Within the inspected set, semantic mismatch accounts for \textbf{50.0\%}: the prediction-derived text prototype emphasizes cues that are weak, misleading, or absent in the image. The remaining cases are categorized as cluttered or poorly localized regions (\textbf{21.4\%}), other (\textbf{21.4\%}), and visually ambiguous classes (\textbf{7.1\%}).

Representative examples further illustrate these patterns. Semantic mismatch often occurs when the selected regions are locally plausible but semantically align with misleading text-side cues rather than the ground-truth class under distribution shift. Cluttered or poorly localized cases arise when the proposal set fails to fully isolate the truly discriminative content at inference time. Visually ambiguous cases remain difficult even after localized refinement because competing classes still share highly similar local attributes in practice. The remaining ''other'' cases include failures that do not fall cleanly into a single dominant category. Overall, these observations suggest that future improvements may come from stronger proposal generation quality and more robust text prototype construction.

\section{Stage~2-only Failure Cases}
\label{app:stage2_only_cases}

To illustrate the Stage~2-only ablation in \Cref{fig:ablation_effectiveness}, we show two representative failures. This variant removes class-agnostic visual initialization and applies semantic refinement from the beginning, allowing uncertain early semantic estimates to influence region selection before stable object-centric candidates are established. These examples illustrate a failure mode but do not provide independent quantitative evidence.

\begin{figure*}[t]
    \centering
    \includegraphics[width=\linewidth]{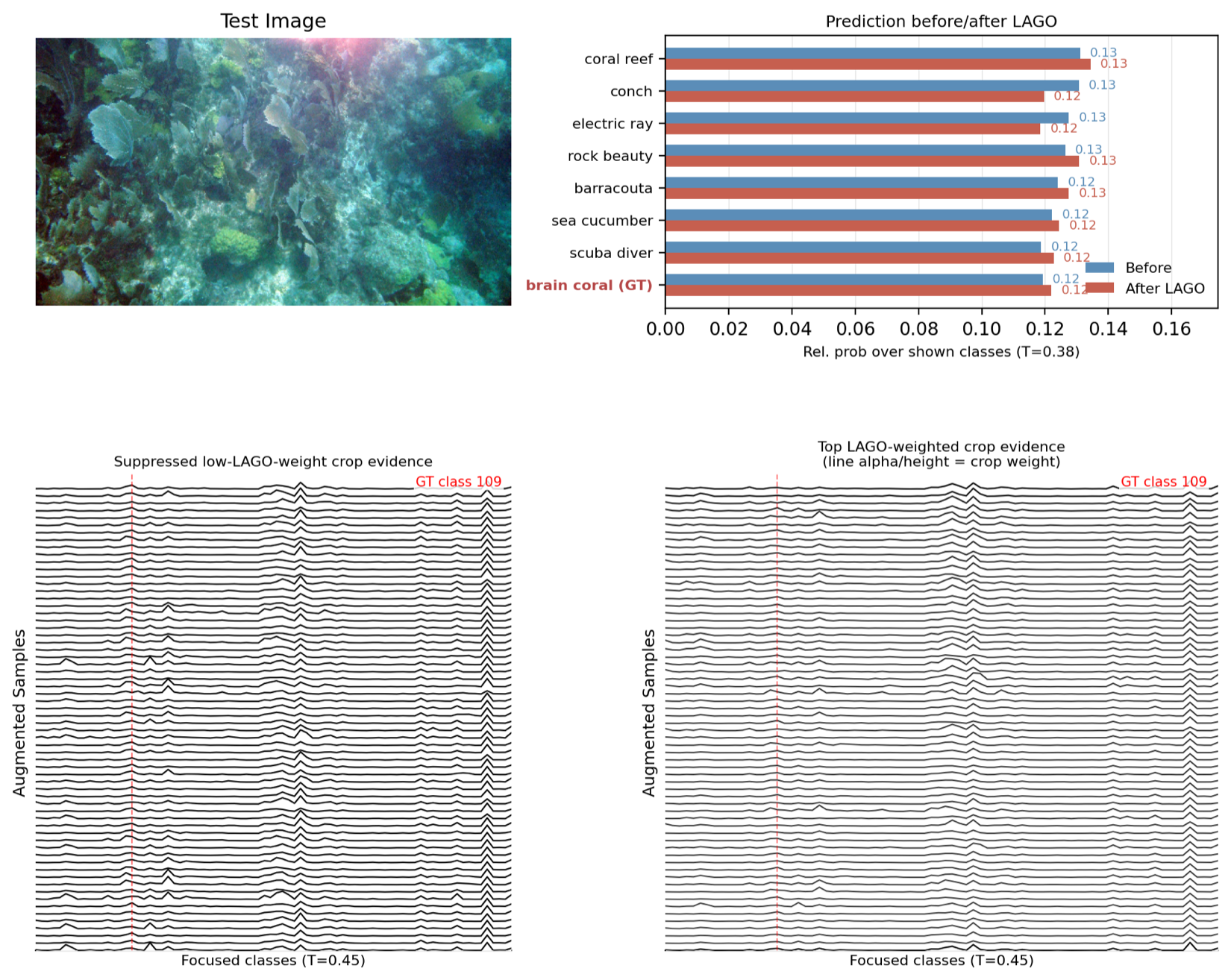}
    \caption{\textbf{Stage~2-only failure case on an underwater scene.}
    The ground-truth class is \textit{brain coral}. Without class-agnostic visual initialization, Stage~2-only refinement produces diffuse evidence across related underwater categories. The ground-truth class is not separated after refinement, illustrating how unreliable early semantics can bias region selection and reinforce ambiguous predictions.}
    \label{fig:stage2_only_case_brain_coral}
\end{figure*}

\begin{figure*}[t]
    \centering
    \includegraphics[width=\linewidth]{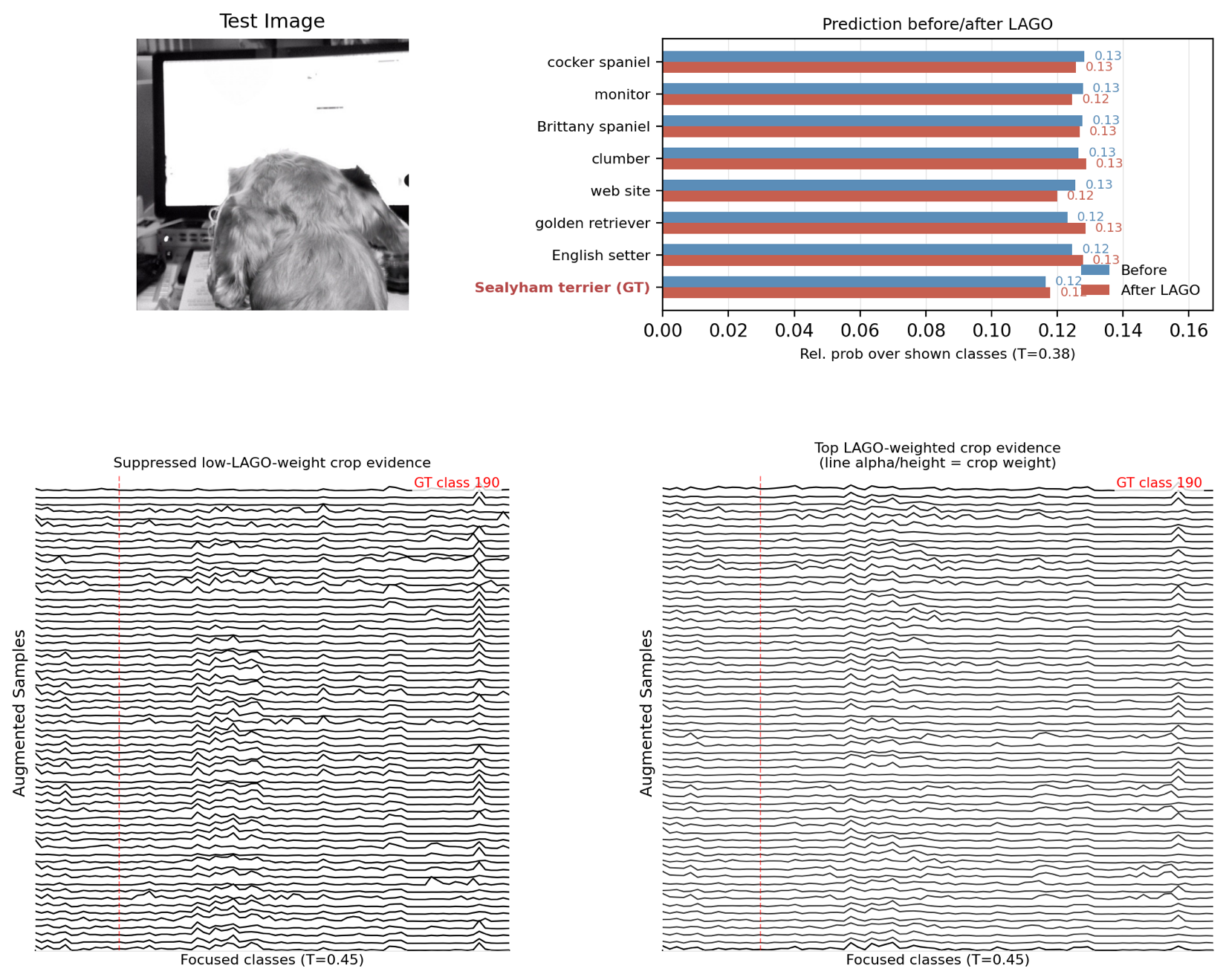}
    \caption{\textbf{Stage~2-only failure case on a fine-grained pet image.}
    The ground-truth class is \textit{Sealyham terrier}. Without stable class-agnostic initialization, Stage~2-only refinement remains diffuse over related dog categories and confounding classes. This shows that applying semantic refinement from the start can overcommit to unreliable early evidence rather than isolate the discriminative region.}
    \label{fig:stage2_only_case_sealyham}
\end{figure*}

As shown in \Cref{fig:stage2_only_case_brain_coral,fig:stage2_only_case_sealyham}, Stage~2-only refinement does not reliably separate the ground-truth class from visually or semantically related alternatives. In these examples, uncertain early hypotheses bias region selection toward evidence that preserves an incorrect or ambiguous prediction. Full LAGO reduces this risk by establishing class-agnostic object-centric candidates before applying confidence-aware text-guided refinement.

\section{Additional Qualitative Analysis of Crop Reweighting}
\label{app:qualitative_reweighting}

We provide additional qualitative examples to illustrate how LAGO reweights crop-level evidence at inference time. Rather than modifying the text encoder itself, LAGO refines prediction by assigning higher weights to crops that provide more reliable, semantically discriminative object-level evidence and suppressing crops that are less informative, ambiguous, or dominated by background or context.

In each example, the top-left panel shows the original test image, and the top-right panel compares the prediction distribution before and after applying LAGO at inference time over a focused set of candidate classes. The bottom-left panel shows the prediction distributions of crops assigned relatively low weights by LAGO, while the bottom-right panel shows the distributions of the top LAGO-weighted crops. For readability, the ridge plots are shown over a visually interpretable focused subset of classes, and the red dashed line marks the ground-truth class. The visualization temperature and candidate subset are strictly used only for display and do not affect the model's final prediction.

\begin{figure*}[t]
    \centering
    \includegraphics[width=\textwidth]{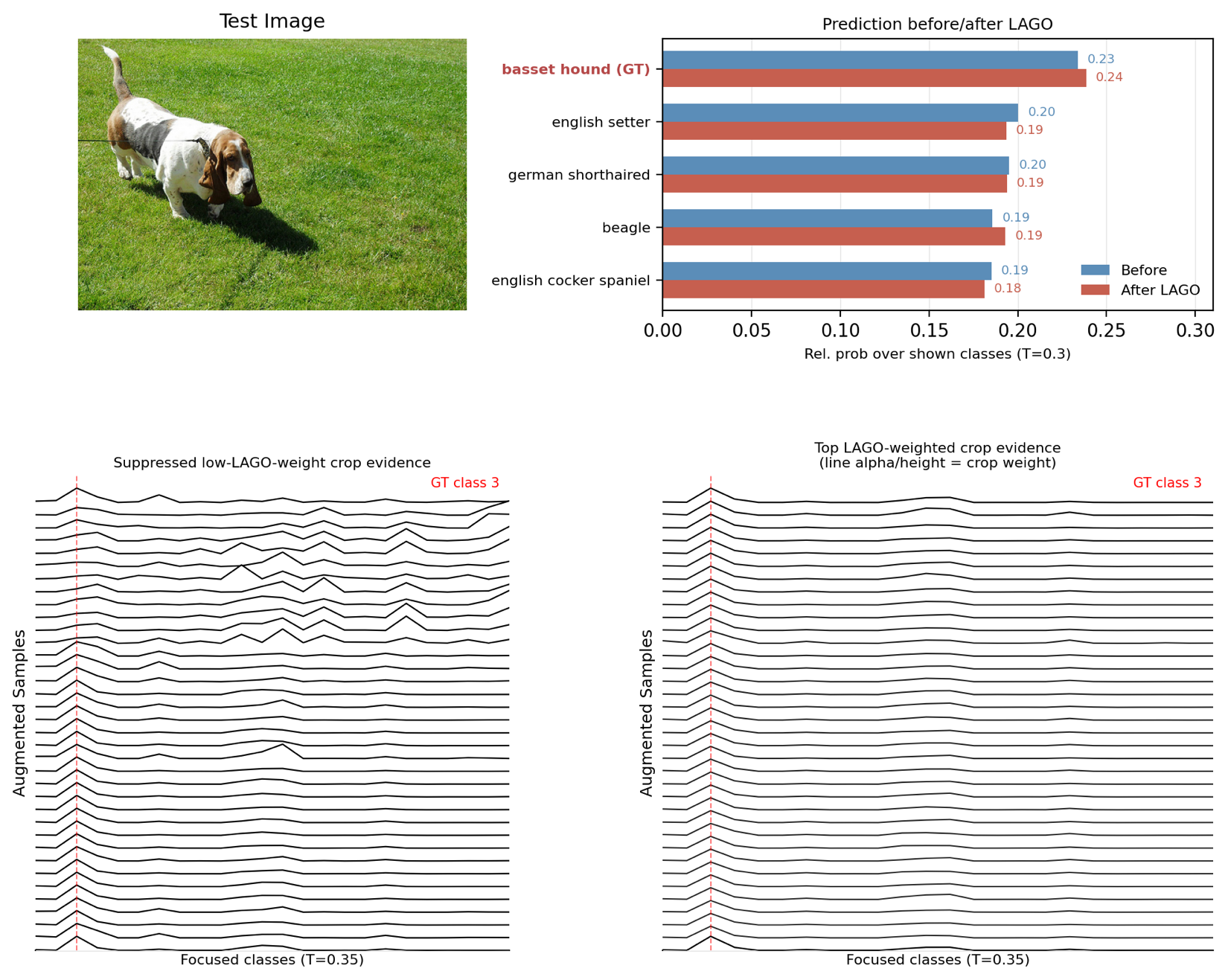}
    \caption{\textbf{Qualitative example of crop reweighting on Oxford Pets (basset hound).}
    In this selected example, low-weight crops contain weaker or less object-focused evidence, while highly weighted crops concentrate more mass near the displayed ground-truth class. The figure visualizes the weighting behavior rather than an aggregate effect.}
    \label{fig:qual_basset}
\end{figure*}

\begin{figure*}[t]
    \centering
    \includegraphics[width=\textwidth]{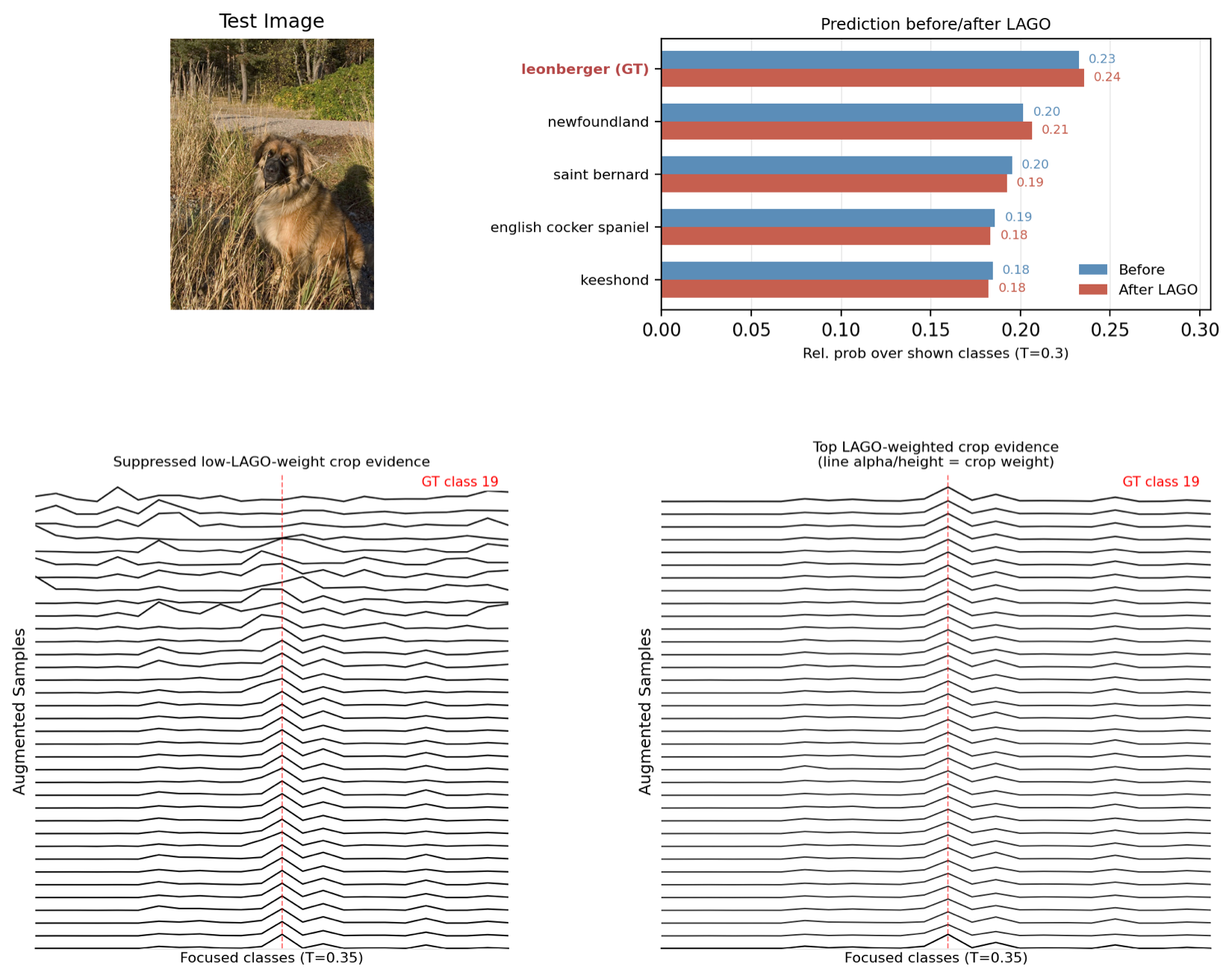}
    \caption{\textbf{Qualitative example of crop reweighting on Oxford Pets (leonberger).}
    In this selected example, the highly weighted crops show more concentrated evidence near the displayed target than the low-weight crops.}
    \label{fig:qual_leonberger}
\end{figure*}

\begin{figure*}[t]
    \centering
    \includegraphics[width=\textwidth]{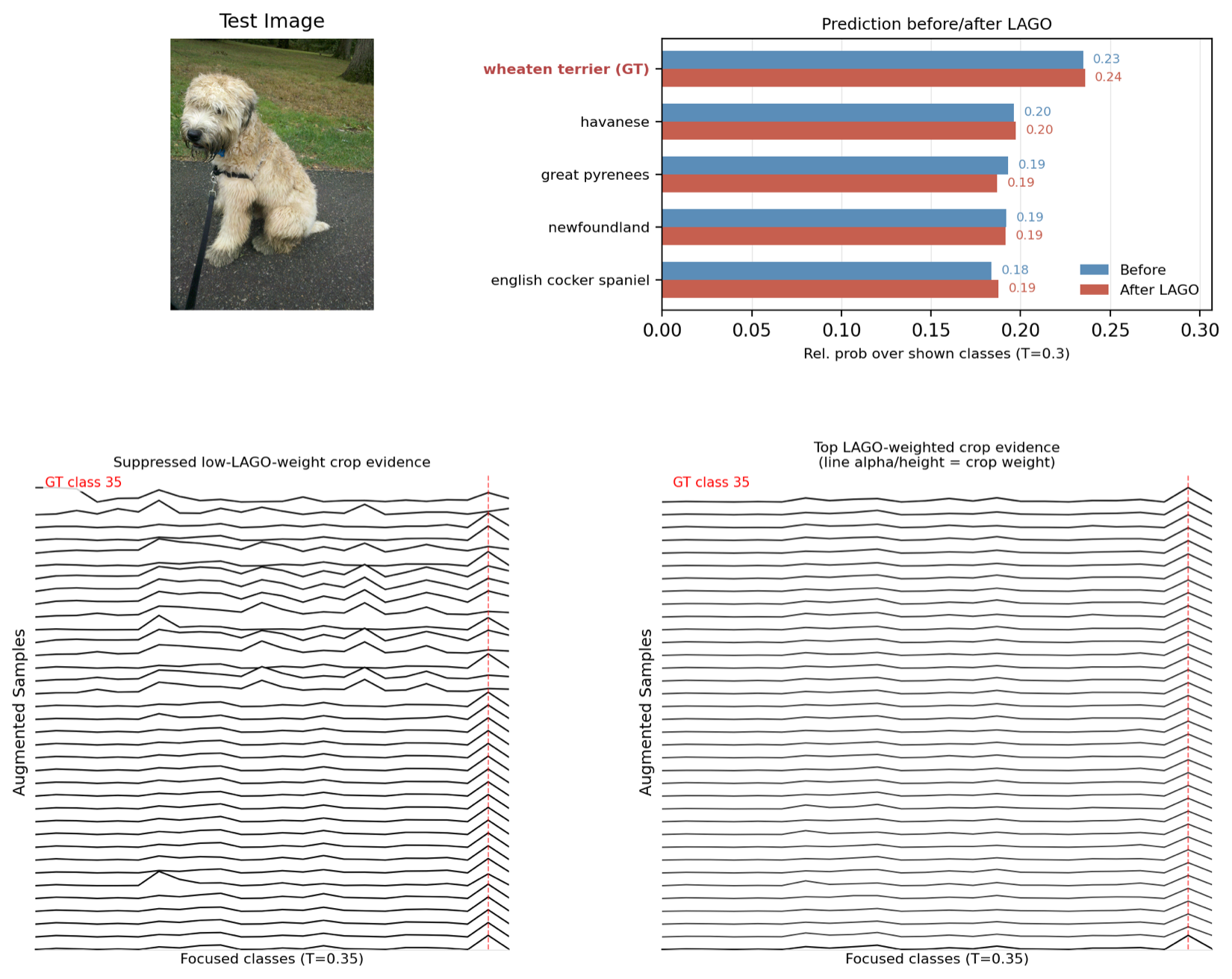}
    \caption{\textbf{Qualitative example of crop reweighting on Oxford Pets (wheaten terrier).}
    In this selected example, the aggregated distribution places more emphasis on object-focused crops and becomes more concentrated around the displayed ground-truth class.}
    \label{fig:qual_wheaten}
\end{figure*}

\begin{figure*}[t]
    \centering
    \includegraphics[width=\textwidth]{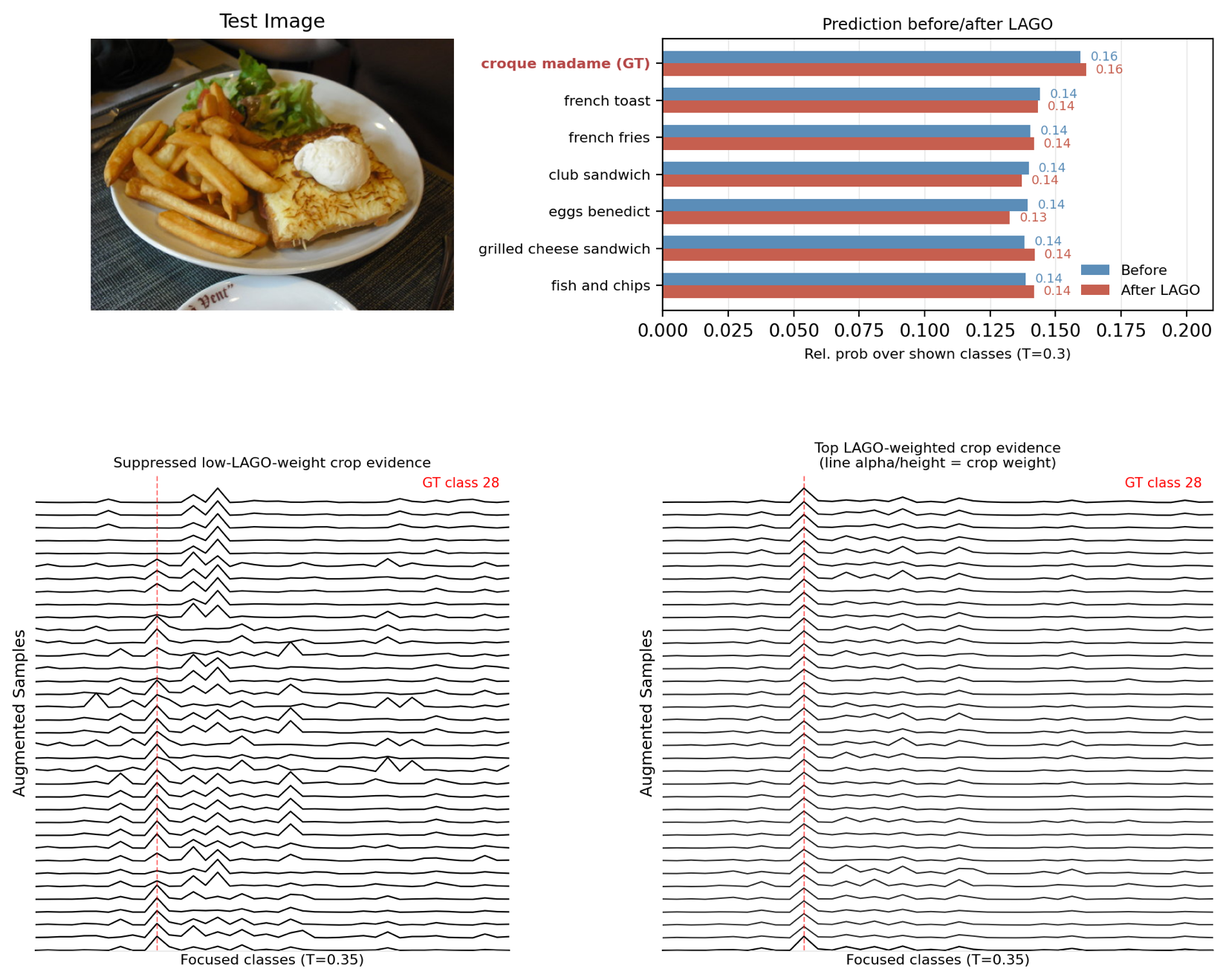}
    \caption{\textbf{Qualitative example of crop reweighting on Food101 (croque madame).}
    In this selected example, highly weighted crops capture more of the displayed food structure than low-weight crops, and the aggregated distribution favors the ground-truth class.}
    \label{fig:qual_food}
\end{figure*}

\begin{figure*}[t]
    \centering
    \includegraphics[width=\textwidth]{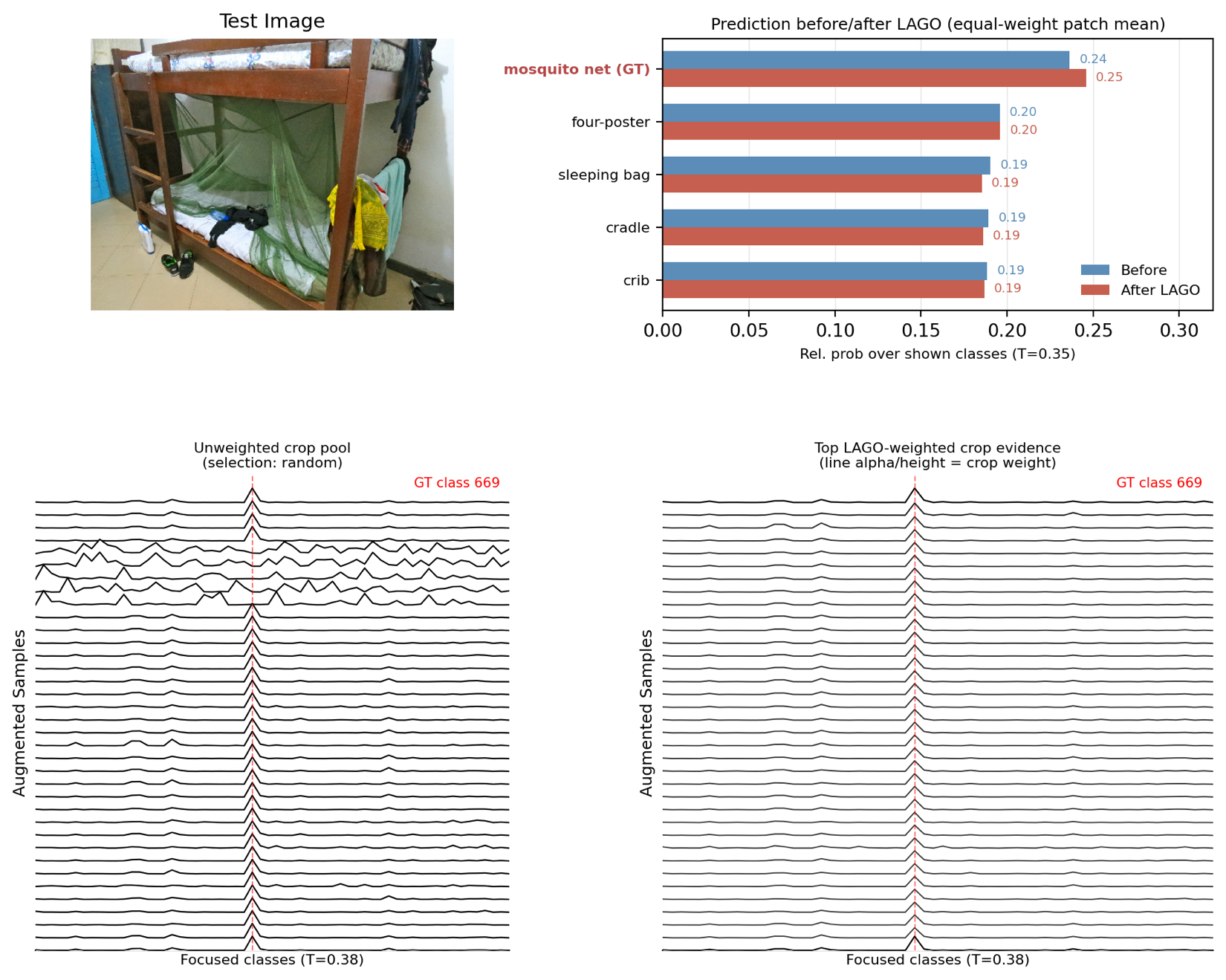}
    \caption{\textbf{Qualitative example of crop reweighting on ImageNet-V2 (mosquito net).}
    In this selected example, the weighted crop evidence is more concentrated around the displayed ground-truth class after aggregation.}
    \label{fig:qual_imagenetv2}
\end{figure*}

\begin{figure*}[t]
    \centering
    \includegraphics[width=\textwidth]{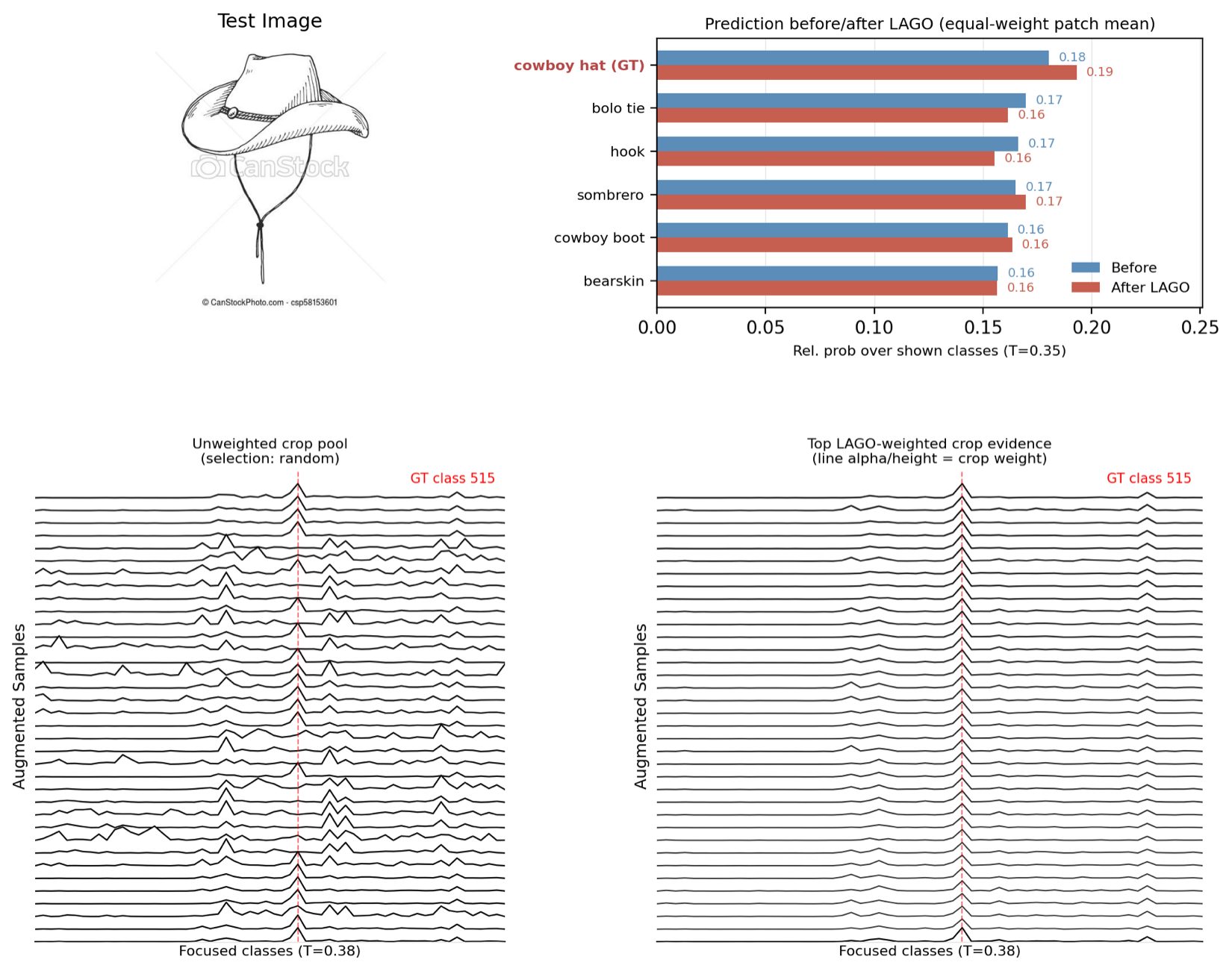}
    \caption{\textbf{Qualitative example of crop reweighting on ImageNet-R (cowboy hat).}
    In this selected stylized example, the highly weighted crops are more consistent with the displayed target object, and the final distribution separates the ground-truth class more clearly.}
    \label{fig:qual_imagenetr}
\end{figure*}

Across the selected examples, LAGO downweights background-dominated or semantically diffuse crops and emphasizes crops whose evidence is more concentrated around the displayed object. The resulting distributions visualize the intended aggregation behavior; because these are selected cases over a display-only class subset, they should not be interpreted as an independent performance estimate. Oxford Pets examples appear in Figures~\ref{fig:qual_basset}--\ref{fig:qual_wheaten}, with additional Food101, ImageNet-V2, and ImageNet-R cases in Figures~\ref{fig:qual_food}--\ref{fig:qual_imagenetr}.

\section{Additional Human Evaluation Analysis}
\label{app:human_eval}

We provide an exploratory human evaluation of whether Stage~2 region paths align with human-perceived discriminative evidence. This analysis complements accuracy by asking whether selected regions contain evidence that annotators consider useful for recognizing the ground-truth class. It is not used for model selection or tuning and is not intended as a population-level user study.

We conduct the study on a selected set of 100 samples spanning CUB, Oxford Pets, ImageNet, DTD, and ImageNet-R. Because the sample set is curated rather than randomly drawn to estimate population rates, all percentages below describe this study set only. Four independent evaluators annotate the same questionnaire. For each image, they are shown the original image, the ground-truth class label, and two region sets generated by Stage~1 and Stage~2. The pairwise options are anonymized and randomized. Evaluators answer: (1) which option better supports recognition, and (2) whether the Stage~2 regions capture key evidence for the ground-truth class.

\paragraph{Evaluation protocol.}
The first question provides a \emph{more direct} relative comparison between Stage~1 and Stage~2:
\emph{Which region set better supports recognizing the ground-truth class?}
The displayed answer is one of the two anonymized options or a tie. After unblinding with the hidden option mapping, we convert each response to one of {\texttt{Stage1}, \texttt{Stage2}, \texttt{Tie}}. This relative comparison assesses annotator preference for the complete Stage~2 refinement over visual-only Stage~1; it does not isolate confidence modulation.

The second question provides an \emph{absolute} human judgment of Stage~2:
\emph{Do the Stage~2 regions capture the key visual evidence needed to recognize the ground-truth class?}
The answer is one of {\texttt{Yes}, \texttt{Partially}, \texttt{No}}. We report both a strict binary alignment rate, where only \texttt{Yes} is counted as aligned, and a soft partial-credit score, where \texttt{Yes}/\texttt{Partially}/\texttt{No} are mapped to $1/0.5/0$, respectively.

\subsection{Overall Results}

Table~\ref{tab:human_eval_summary} summarizes the overall results from human evaluation under this protocol.
Across the four evaluators, Stage~2 is preferred in $263/400$ responses ($65.8\%$), increasing to $263/338$ ($77.8\%$) after ties are excluded. Evaluator D is more conservative in the pairwise comparison, preferring Stage~2 in $46/100$ responses, or $46/78$ after excluding ties. At the item level, all four evaluators prefer Stage~2 on $32/100$ samples, a strict majority do so on $57/100$, at least half on $78/100$, and at least one on $96/100$. Comparing Stage~2 and Stage~1 votes per sample, Stage~2 receives more votes on 75 samples, the stages tie on 13, and Stage~1 receives more on 12. These results show a pooled directional preference for Stage~2 on the selected study set, while the evaluator-level variation and agreement statistics below limit stronger conclusions.

For the absolute Stage~2 judgment, the pooled strict alignment rate is $263/400$ ($65.8\%$), with a partial-credit mean of $0.766$. Evaluator D assigns high absolute scores ($82.0\%$ strict alignment and $0.900$ partial credit) despite a more conservative pairwise preference, illustrating that absolute plausibility and relative improvement are different questions. Overall, Stage~2 regions are often judged plausible on this study set, but the judgment is subjective and evaluator-dependent.

\begin{table}[!t]
\centering
\captionsetup{skip=8pt}
\caption{\textbf{Exploratory human evaluation on the selected 100-image study set.} Results show pooled Stage~2 preference and absolute plausibility judgments; they vary across evaluators and do not isolate confidence modulation.}
\label{tab:human_eval_summary}
\small
\setlength{\tabcolsep}{7pt}
\begin{tabular}{lccccc}
\toprule
\textbf{Scope} & \textbf{Answered} & \textbf{Stage~2 Pref.} & \textbf{Pref. (w/o Ties)} & \textbf{Strict Align.} & \textbf{Partial Score} \\
\midrule
Evaluator A & 100 & 82.0\% & 83.7\% & 74.0\% & 0.865 \\
Evaluator B & 100 & 70.0\% & 89.7\% & 48.0\% & 0.585 \\
Evaluator C & 100 & 65.0\% & 77.4\% & 59.0\% & 0.715 \\
Evaluator D & 100 & 46.0\% & 59.0\% & 82.0\% & 0.900 \\
\midrule
\textbf{Pooled} & \textbf{400} & \textbf{65.8\%} & \textbf{77.8\%} & \textbf{65.8\%} & \textbf{0.766} \\
\bottomrule
\end{tabular}
\end{table}

\subsection{Agreement Between Evaluators}

We examine inter-rater agreement using Fleiss' $\kappa$ across the four evaluators. For Question~1, the mean per-item observed agreement is $0.627$. On anonymized option labels (\texttt{A}/\texttt{B}/\texttt{Tie}), Fleiss' $\kappa=0.390$; after mapping to semantic labels (\texttt{Stage1}/\texttt{Stage2}/\texttt{Tie}), observed agreement remains $0.627$ but $\kappa$ drops to $0.266$ because the marginals favor Stage~2. Pairwise Cohen's $\kappa$ ranges from $0.046$ to $0.434$. Thus the pooled preference does not imply strong item-level consensus about which particular samples improve.

Agreement is lower for absolute Stage~2 alignment: the three-way label (\texttt{Yes}/\texttt{Partially}/\texttt{No}) yields Fleiss' $\kappa=0.108$, and its strict binary version yields $\kappa=0.160$. We therefore treat the study as exploratory evidence of a pooled directional preference and frequent perceived plausibility, not as evidence of high rater reliability or population-level effect sizes.

\subsection{Dataset-wise Trends}

\begin{figure}[!htbp]
    \centering
    \includegraphics[width=0.82\linewidth]{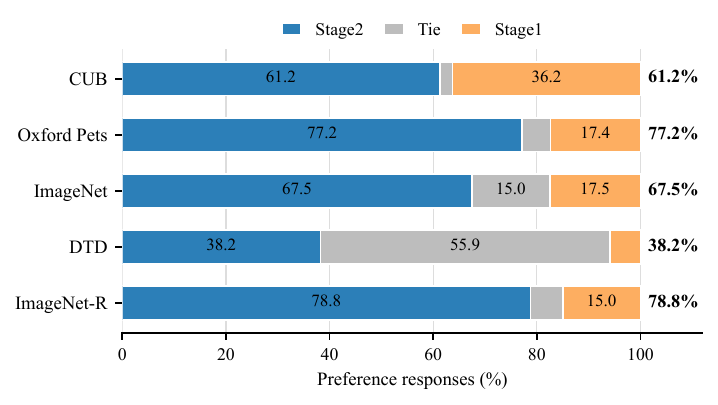}
    \caption{\textbf{Dataset-wise relative preference distribution over Stage~2, tie, and Stage~1.} 
    Stage~2 is strongest on ImageNet-R and Oxford Pets, followed by ImageNet and CUB, while DTD shows many ties and remains favorable to Stage~2 among non-tie responses under the human evaluation protocol.}
    \label{fig:human_eval_stage2_pref_by_dataset}
\end{figure}

\begin{figure}[!htbp]
    \centering
    \includegraphics[width=0.82\linewidth]{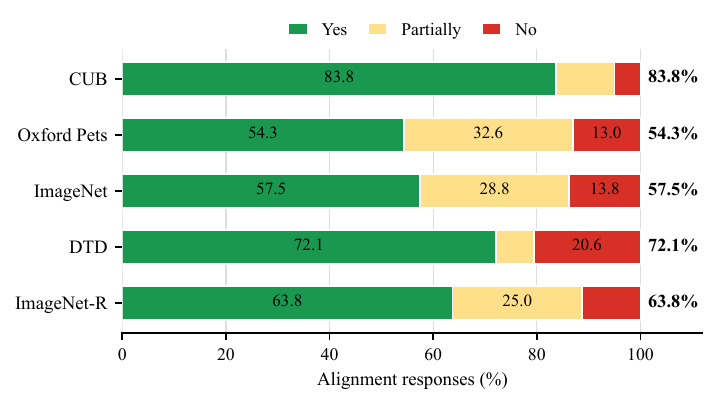}
    \caption{\textbf{Dataset-wise absolute Stage~2 alignment distribution over \texttt{Yes}, \texttt{Partially}, and \texttt{No}.} 
    The strict alignment rate is highest on CUB, followed by DTD and ImageNet-R, while ImageNet and Oxford Pets are lower under the strict criterion in the human evaluation setting used here.}
    \label{fig:human_eval_alignment_by_dataset}
\end{figure}

Figures~\ref{fig:human_eval_stage2_pref_by_dataset} and~\ref{fig:human_eval_alignment_by_dataset} break down the human evaluation results by dataset.
Within this selected study set, Stage~2 preference is highest on ImageNet-R and Oxford Pets ($78.8\%$ and $77.2\%$), followed by ImageNet ($67.5\%$) and CUB ($61.3\%$). DTD is tie-heavy: Stage~2 is selected in $26/68$ responses when ties are included and in $26/30$ non-tie responses. These descriptive breakdowns are not accompanied by dataset-wise uncertainty tests and should not be read as a ranking of population-level effectiveness.

Strict alignment rates on the selected samples are $83.8\%$ for CUB, $72.1\%$ for DTD, $63.7\%$ for ImageNet-R, $57.5\%$ for ImageNet, and $54.3\%$ for Oxford Pets. These values describe perceived path plausibility under the present annotation protocol; they neither measure classification improvement nor establish statistically reliable cross-dataset differences.

\subsection{Relation to Prediction Correctness}

We also compare perceived path quality with final prediction correctness. Under the pooled strict criterion, alignment is $184/284=64.8\%$ for correct predictions and $79/116=68.1\%$ for incorrect predictions; partial-credit scores are $0.764$ and $0.772$. These close descriptive rates show that perceived path plausibility does not trivially coincide with correctness on this selected set. They do not establish independence or a negative association, because dataset composition and difficulty are uncontrolled.

\begin{figure}[!htbp]
    \centering
    \includegraphics[width=0.95\linewidth]{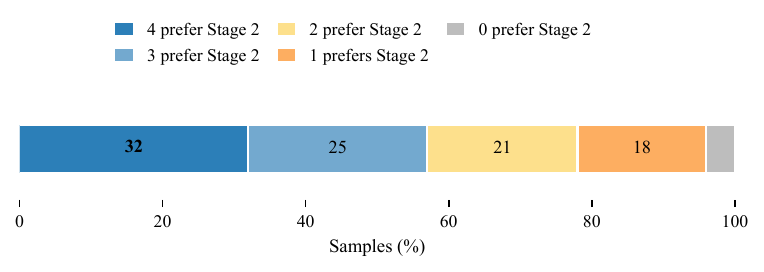}
    \caption{\textbf{Item-level support for Stage~2 preference across the four evaluators.} 
    All four evaluators prefer Stage~2 on $32/100$ samples, a strict majority do so on $57/100$, at least half on $78/100$, and at least one on $96/100$. These counts describe the selected study set and coexist with the limited agreement statistics reported in Appendix~\ref{app:human_eval}.}
    \label{fig:human_eval_preference_consensus}
\end{figure}

\begin{figure}[!htbp]
    \centering
    \includegraphics[width=0.82\linewidth]{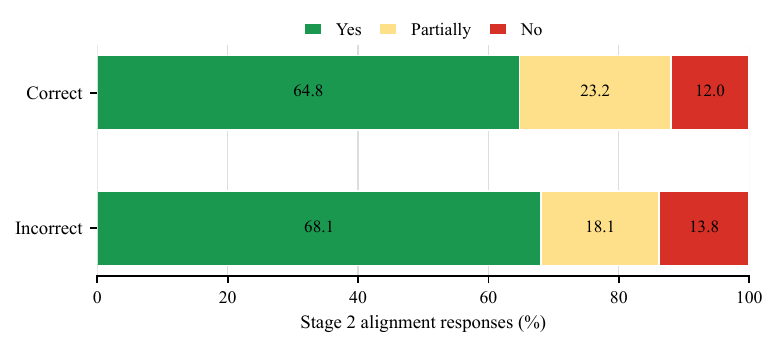}
    \caption{\textbf{Stage~2 alignment distribution conditioned on whether the final prediction is correct.}
    In this subset, human-perceived path plausibility does not simply track final decision correctness, especially in fine-grained or semantically ambiguous recognition settings at inference time.}
    \label{fig:human_eval_correctness_alignment}
\end{figure}

Figures~\ref{fig:human_eval_preference_consensus} and~\ref{fig:human_eval_correctness_alignment} summarize item-level preference and alignment conditioned on correctness. Visually plausible paths also occur among incorrect predictions, so a plausible localization is not sufficient for a correct decision. Fine-grained ambiguity, text-side mismatch, or final aggregation may still produce errors; the annotation study does not attribute causes among these possibilities.

\subsection{Qualitative Implications}

\begin{figure}[!htbp]
    \centering
    \includegraphics[width=0.98\linewidth]{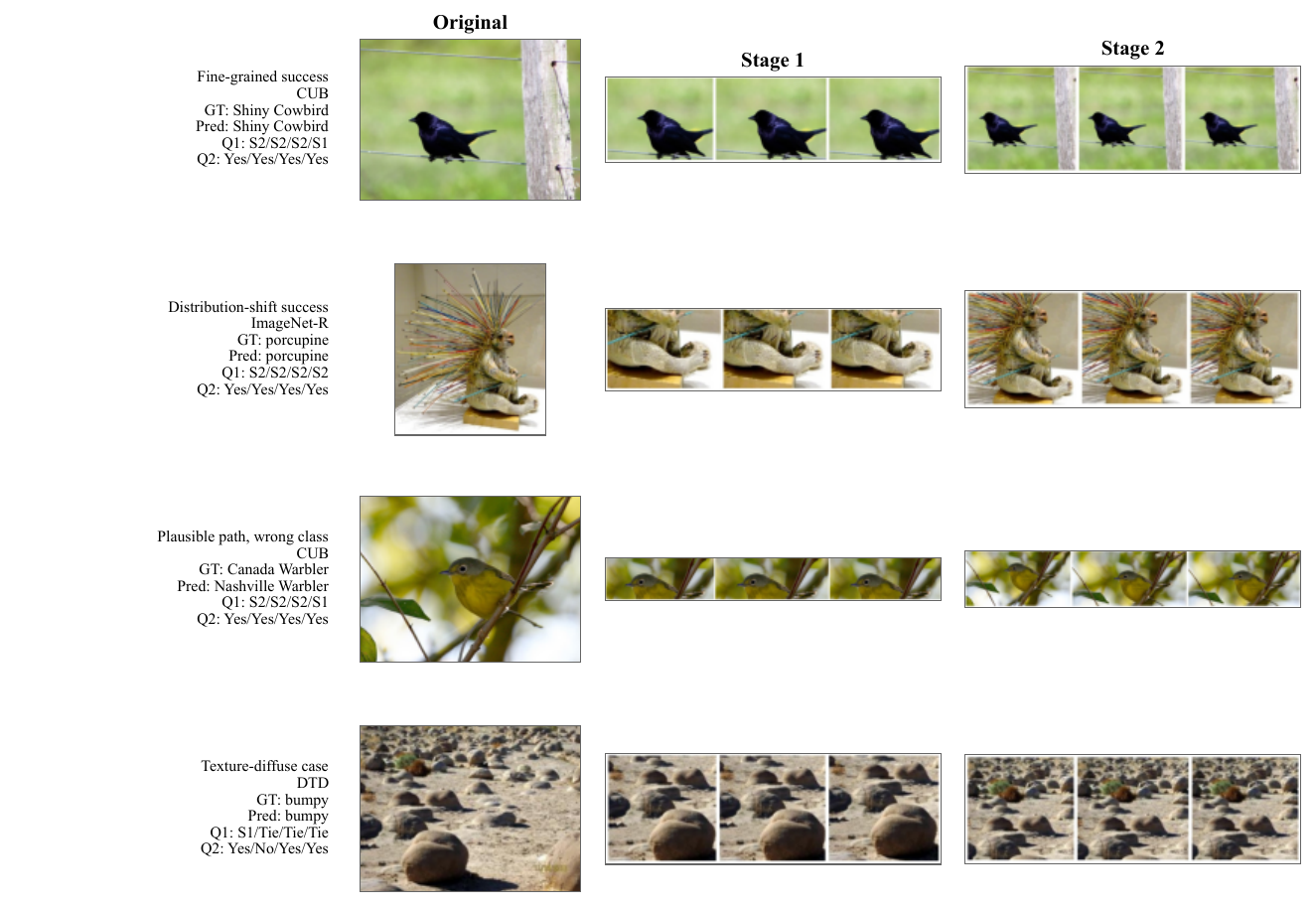}
    \caption{\textbf{Representative human evaluation examples.} 
    Each row shows the image, Stage~1 regions, and Stage~2 regions, together with the ground-truth label, model prediction, and evaluator votes.}
    \label{fig:human_eval_qualitative_examples}
\end{figure}

Figure~\ref{fig:human_eval_qualitative_examples} shows representative examples from the human evaluation across datasets. The qualitative cases illustrate the same two main conclusions as the quantitative results. First, Stage~2 often produces more semantically focused, discriminative, and visually grounded region sets than Stage~1, consistent with its strong non-tie human preference rate and majority item-level support. Second, a visually plausible Stage~2 path does not always imply a correct final prediction at inference time, especially in fine-grained recognition where nearby categories can still share highly similar local evidence.

Overall, the human evaluation provides qualitative evidence about region-path plausibility. Stage~2 is favored over Stage~1 after excluding ties, which supports the ordered transition from visual initialization to semantic refinement but does not isolate the confidence controller or establish an accuracy gain. Human-aligned paths also do not always produce correct final predictions, highlighting the distinction between \emph{path plausibility} and \emph{decision correctness}. This distinction is especially relevant in fine-grained recognition, where a region may capture a meaningful attribute while remaining insufficient to resolve nearby categories.

\subsection{Participant Instructions and Compensation}
\label{app:human_eval_instructions}

The human evaluation was conducted as an offline annotation study using static visual questionnaires. Each questionnaire item contained the original image, the ground-truth class label, and two anonymized candidate region sets corresponding to Stage~1 and Stage~2, shown as Option A and Option B in randomized order. Evaluators were not told which option corresponded to which stage during annotation. The evaluation was used only for qualitative and quantitative analysis of region-path plausibility and was not used for model selection, hyperparameter tuning, or training.

The full participant-facing instruction text below was presented identically to all evaluators during the complete annotation study and is reproduced verbatim for transparent review:

\begingroup
\begin{quote}
You will be shown an image, its ground-truth class label, and two anonymized sets of candidate regions. Please judge the region sets based on whether they contain visual evidence useful for recognizing the ground-truth class. Do not infer which method produced each option. For each sample, answer the following two questions.
Question 1: Which region set better supports recognizing the ground-truth class? Please choose one of: Option A, Option B, or Tie.
Question 2: Do the Stage~2 regions capture the key visual evidence needed to recognize the ground-truth class? Please choose one of: Yes, Partially, or No.
\end{quote}
\endgroup

For Question~1, the mapping between Option A/Option B and Stage~1/Stage~2 was hidden from evaluators and was used only after annotation for analysis. For Question~2, evaluators were shown the Stage~2 region set and asked to judge its absolute adequacy for recognizing the ground-truth class. No additional personal information was collected from the evaluators. The questionnaire consisted only of static images, class labels, candidate region visualizations, and multiple-choice responses.

No crowdworking platform was used. The four evaluators participated in an internal offline evaluation and completed the same questionnaire independently. No monetary compensation was provided for this voluntary, low-burden exploratory analysis. The study collected only the task responses described above and no identifying, demographic, or sensitive personal information; we therefore consider the expected risk minimal.

\subsection{Participant Risk Disclosure and Ethics Review}
\label{app:human_eval_irb}

The human evaluation involved non-sensitive visual judgments over public benchmark images. Evaluators were informed that the study assessed whether candidate region sets provide useful evidence for recognizing a supplied ground-truth label. The study involved no deception or intervention and collected only multiple-choice task responses, without identifying, demographic, or other sensitive information. The expected risk was therefore minimal.

Under the applicable institutional policy, this exploratory internal annotation study did not require formal IRB approval or equivalent ethics-board review, because it involved minimal-risk non-sensitive visual judgments. Participation was voluntary, and evaluators could stop the annotation at any time. No identifying information about the evaluators is included in the paper or supplemental material.

\section{Additional Discussion}

Fine-grained class descriptions become useful when the model can retrieve the visual evidence they describe. LAGO operationalizes this requirement as directed region discovery: descriptions contribute to the selection of image regions, rather than serving only as endpoints for similarity computation. This view connects text quality and localization directly. A part or attribute description can influence the decision when the search exposes the corresponding evidence and the aggregation preserves it.

The experiments support an ordered solution to this problem. Class-agnostic object-centric initialization supplies coverage and diversity before a semantic hypothesis is available. The prediction-derived text prototype then directs refinement toward class-relevant evidence, and object-context aggregation restores information that tight crops may discard. The controlled CUB mechanism experiment separates directed search from post-hoc semantic reranking: language changes 786/1,000 crop sets, improves query relevance, and yields a significant $+0.80$-point gain on that subset. Candidate-budget results further show that the benefit emerges after adequate proposal coverage. By contrast, the fixed component audit provides weaker support for the particular confidence mapping, whose independent accuracy gain is not significant after correction.

The protocol controls also define the scope of the empirical claim. Dataset-specific evaluation-label calibration shows attainable post-hoc performance and is reported only as a diagnostic. With a single ImageNet-V2-specified tuple frozen before target evaluation, LAGO improves numerically over full-image inference on all eight transfer datasets. Against WCA, it is higher on four representative matched targets, and three differences remain significant after Holm correction. The additional two-backbone DTD audit is positive on both backbones but nonsignificant. These results support broad transfer and effective candidate use without implying universal dominance across every dataset--backbone pair. Broader fixed-protocol evaluation is needed to determine when confidence modulation adds a reliable aggregate benefit.

\section{Broader Impacts}
\label{app:broader_impacts}

LAGO aims to improve zero-shot visual recognition by making visual-text alignment more localized, selective, and candidate-efficient. Potential positive impacts include reducing the need for task-specific model training and lowering view redundancy relative to random-crop alignment under matched candidate budgets. These properties may help in settings where annotation is expensive or models must generalize to unseen categories, although the substantially higher cold construction latency limits deployment benefits for ordinary one-pass workloads.

At the same time, more capable zero-shot recognition can introduce negative impacts when adopted without governance and auditing. Visual recognition systems may be used in surveillance, profiling, or other applications where individuals or groups are monitored without appropriate consent. Because LAGO relies on pretrained vision-language models and offline text descriptions, it may inherit biases from the underlying model, datasets, or textual descriptions, potentially producing uneven performance across visual domains, object categories, cultural contexts, or demographic groups.

The method is intended as a research contribution for evaluating and improving localized visual-text alignment, not as a standalone deployed decision-making system. Any deployment in sensitive domains should include additional evaluation for bias, privacy, robustness, and domain-specific failure modes. In particular, users should validate performance on the target population and application setting, avoid using the method for high-stakes decisions without human oversight and accountability, and ensure that data collection and use comply with consent, privacy, and legal requirements.

\end{document}